%% file: main.tex
\documentclass[11pt]{article}

\usepackage[final]{acl}

\usepackage{times}
\usepackage{latexsym}
\usepackage[T1]{fontenc}
\usepackage[utf8]{inputenc}
\usepackage{silence}
\usepackage[nopatch]{microtype}
\usepackage{inconsolata}
\usepackage{graphicx}
\usepackage{amsmath}
\usepackage{amssymb}
\usepackage{amsthm}
\usepackage{booktabs}
\usepackage{multirow}
\usepackage{xcolor}
\usepackage{colortbl}
\usepackage{subcaption}
\usepackage{enumitem}
\usepackage[most]{tcolorbox}
\tcbuselibrary{listings}
\usepackage{placeins}
\graphicspath{{figures/}}

\newtheorem{definition}{Definition}

\title{Beyond State Consistency: \\Behavior Consistency in Text-Based World Models}

\author{
  Youling Huang\textsuperscript{1,4,*} \quad
  Guanqiao Chen\textsuperscript{1} \quad
  Junchi Yao\textsuperscript{2} \quad
  Lu Wang\textsuperscript{4,$\dagger$} \quad
  Fangkai Yang\textsuperscript{4} \quad
  Chao Du\textsuperscript{4} \\
  \bfseries
  ChenZhuo Zhao\textsuperscript{3,4,*} \quad
  Pu Zhao\textsuperscript{4} \quad
  Qingwei Lin\textsuperscript{4} \quad
  Saravan Rajmohan\textsuperscript{4} \quad
  Dongmei Zhang\textsuperscript{4} \\[2pt]
  \textsuperscript{1}Dalian University of Technology \quad
  \textsuperscript{2}MBZUAI \quad
  \textsuperscript{3}Peking University \quad
  \textsuperscript{4}Microsoft
}

\begin{document}
\maketitle

\renewcommand{\thefootnote}{\fnsymbol{footnote}}
\footnotetext[1]{Work done during an internship at Microsoft.}
\footnotetext[2]{Corresponding author.}
\renewcommand{\thefootnote}{\arabic{footnote}}

\input{sections/00_abstract}

\input{sections/01_introduction}
\input{sections/02_related_work}

\input{sections/03_formulation}

\input{sections/04_method}
\input{sections/05_experiments}
\input{sections/06_applications}
\input{sections/08_conclusion}

\bibliography{references}

\clearpage
\input{sections/09_appendix}

\end{document}

%% file: sections/00_abstract.tex
%

\begin{abstract}
World models are emerging as critical components for assessing the consequences of actions taken by interactive agents in online planning and offline evaluation. In text-based environments, world models are typically trained and evaluated with single-step metrics such as Exact Match, aiming to improve the similarity between predicted and real-world states, but such metrics have been shown to be insufficient for capturing actual agent behavior. To address this issue, we introduce a new behavior-consistency training paradigm aimed at improving the \emph{functional consistency} between the world model and the real environment. This paradigm focuses on optimizing a tractable step-level reward named \emph{Behavior Consistency Reward} (BehR), which measures how much the likelihood of a logged next action changes between the real state and the world-model-predicted state under a frozen Reference Agent. Experiments on WebShop and TextWorld show that BehR-based training improves task-level functional consistency in several settings, with the clearest gains in WebShop and smaller changes in near-ceiling regimes, while preserving or improving single-step prediction quality in three of four settings. BehR-trained world models also reduce false positives in offline surrogate evaluation and improve inference-time lookahead planning over the corresponding supervised world-model baselines.
Code: \url{https://github.com/Ricardo-H/behr-wm}.
\end{abstract}

%% file: sections/01_introduction.tex
\section{Introduction}
\label{sec:intro}

\begin{figure}[t] 
\centering
\includegraphics[width=0.48\textwidth]{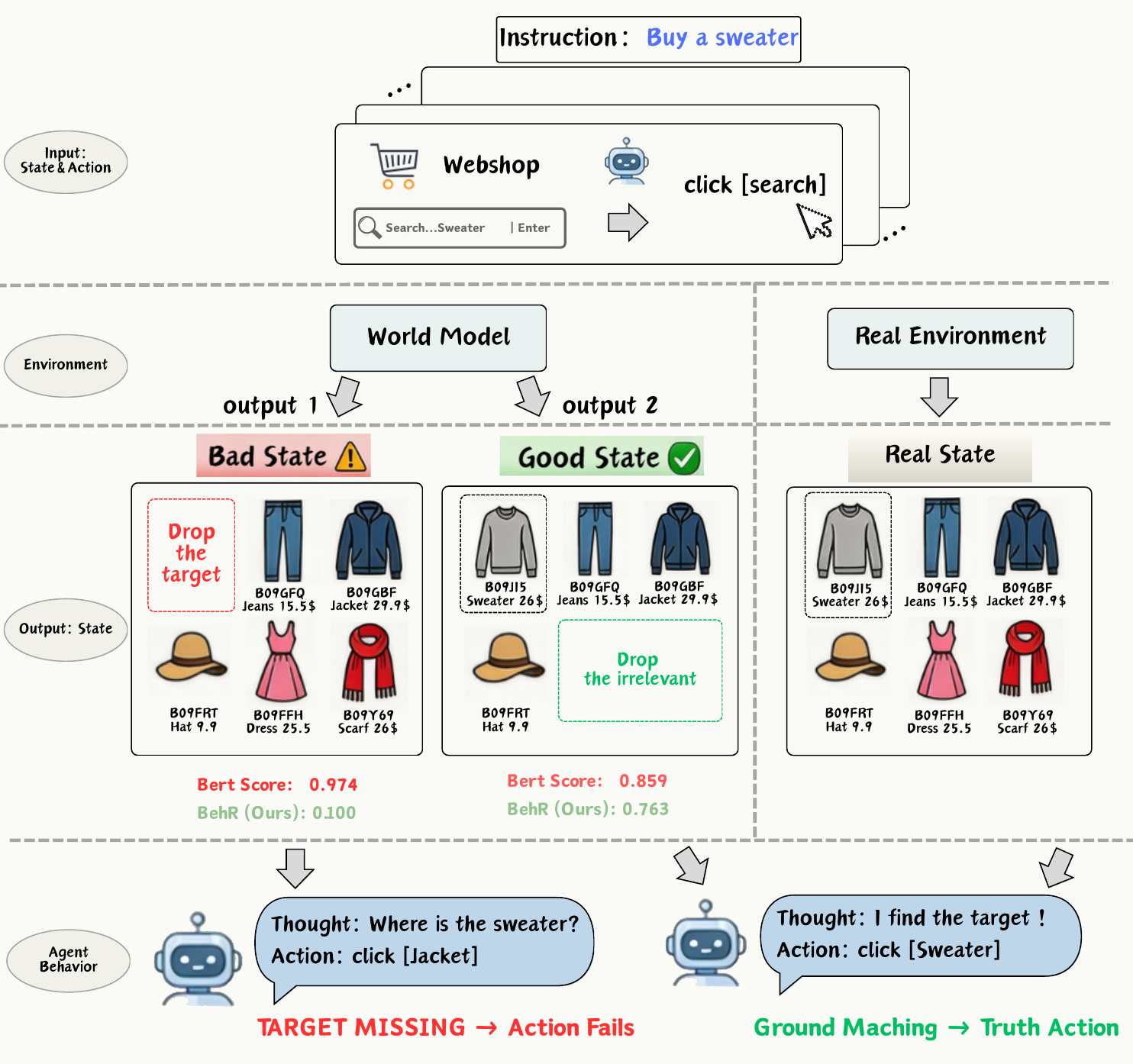}
\caption{\textbf{Metric inversion in WebShop.} The Bad State drops the target product and changes the agent's action despite high surface similarity, while the Good State drops irrelevant products yet preserves the real action. BehR scores this behavioral effect rather than surface overlap.}
\label{fig:metric_inversion}
\end{figure}

Interactive agents deployed in real-world environments such as web navigation, text-based adventure games, and tool-use should be evaluated for their reliability in planning and executing multi-step actions. As testing in real environments is often slow and expensive, \emph{world models} (WMs) are widely used as surrogate environments for both online evaluation and offline benchmarking. Recent work has explored using large language models (LLMs) as world models in text-based settings, including web navigation~\citep{chae2025web} and interactive environments such as WebShop and text adventures~\citep{li2025word}. In these settings, the LLM serves as an environment simulator, generating the next state conditioned on a state-action pair expressed in text.

Most existing training approaches for LLM-based world models focus on \emph{state consistency}: high textual similarity between states predicted by WMs and states observed in the real environment. This training paradigm relies on supervised next-state prediction and reinforcement learning methods to optimize token-level likelihood of the generated states. Still, existing studies have shown that even state-of-the-art SFT-trained WMs with high single-step textual similarity can exhibit substantial task-level inconsistency when agent actions from WM rollouts are replayed in the real environment~\citep{li2025word}. 

As illustrated in Figure~\ref{fig:metric_inversion}, a world model prediction that reproduces most elements on a WebShop page but omits the targeted product can cause the downstream agent to fail. In contrast, a prediction that captures the critical product while missing several irrelevant products may score lower on textual similarity but would still allow the agent to target the right product. As such, metrics measuring state-level consistency may fail at distinguishing good performance from bad performance at the task level, a phenomenon we call \emph{metric inversion}. The appropriate criterion for an LLM-based world model is not ``does the text look similar?'' but rather ``does the agent make the same decision as in the real environment?''


To address this issue, we propose a new training paradigm for LLM-based world models: \textbf{Behavior Consistency Training}. We argue that an ideal world model should strive to achieve \emph{functional consistency}: the predicted state induces the same action distribution as the state from the real environment. Because the full action distribution is typically inaccessible, we formulate \textbf{Behavior Consistency Reward (BehR)} as a tractable step-level proxy rather than an exact surrogate for this ideal objective. Using a frozen Reference Agent, BehR compares the likelihood of the logged next action under the predicted and real states and rewards a logged-action approximation to functional consistency at the step level. Combined with a reinforcement learning method such as Group Relative Policy Optimization (GRPO; \citealt{shao2024deepseekmath}), the resulting \textbf{BehR-WMs} are optimized to generate predictions with high \textbf{Pairwise Consistency Ratio} (CR$_{\text{pw}}$), the fraction of individually successful tasks that remain successful when the action sequence generated in the WMs is replayed in the real environment.

Our work makes three main contributions and evaluates two downstream uses of behavior-consistent WMs.
\begin{enumerate}[leftmargin=*]
    \item \textbf{A behavior-consistent training paradigm for text-based WMs.} We reframe world model learning around functional consistency, propose BehR as a tractable step-level proxy, and formulate CR$_{\text{pw}}$ as the task-level evaluation criterion (\S\ref{sec:method}, \S\ref{sec:setup}).
    \item \textbf{BehR as a promising training signal.} A decision-critical perturbation test shows that BehR correctly captures behavioral severity that surface-level metrics miss, and an optimization-target comparison provides initial evidence that replacing BehR with surface-overlap rewards does not recover the same task-level gains in the tested setting (\S\ref{sec:reward_validation}, \S\ref{sec:reward_ablation}).
    \item \textbf{Improved long-horizon predictive fidelity in selected settings.} In our primary setting, training with BehR substantially improves the Pairwise Consistency Ratio (CR$_{\text{pw}}$) while single-step prediction quality mostly remains stable or improves; gains transfer to other settings but are more heterogeneous, especially in near-ceiling regimes (\S\ref{sec:single_step}, \S\ref{sec:main_results}).
\end{enumerate}
Additionally, BehR-WMs reduce the calibration gap that inflates weak-agent performance during offline evaluation and improve WebShop lookahead planning over the corresponding supervised world-model baselines (\S\ref{sec:surrogate_eval}, \S\ref{sec:lookahead}).

%% file: sections/02_related_work.tex
%
%
%
%

\section{Related Work}
\label{sec:related}

\paragraph{Text-Based World Models.} Traditional world models learn dynamics for planning and control in continuous domains and games \citep{hafner2019dream,schrittwieser2020mastering}; recent work instead uses LLMs as text-based simulators for web and game environments \citep{yao2022webshop,sodhi2023step,zhou2023webarena,wang2024can}. These studies show that language models can generate plausible textual dynamics, but faithful simulator use also requires preserving action-relevant consequences. Our closest baseline is Word2World \citep{li2025word}, which formalizes text-based world models as next-state generators under a standard interaction protocol and trains them with Supervised Fine-Tuning (SFT), but its objective remains token-level likelihood rather than functional consistency.

\paragraph{World-Model Training Objectives and Agent Experience.} Existing text-based world models mainly optimize token-level state consistency \citep{li2025word,chae2025web}. Recent variants such as RLVR-World and CUWM explore RL or agent-centric metrics for world-model improvement~\citep{wu2026rlvr,guan2026computer}, but their rewards or comparisons are still based on surface overlap or discrete action agreement. A complementary line uses learned simulators or synthesized experience to improve the agent policy itself, including DreamGym~\citep{chen2025scaling}, early-experience training~\citep{zhang2025early}, and WebEvolver~\citep{fang2025webevolver}. Our focus is orthogonal: rather than training the agent with synthetic experience, we post-train the simulator so predicted observations better preserve downstream agent behavior. This makes our use of GRPO \citep{shao2024deepseekmath} closer in spirit to alignment than to ordinary next-token fine-tuning, but the target is simulator dynamics rather than human preference alignment \citep{ouyang2022training}.

\paragraph{Evaluation of Simulators.} Text simulators are still often judged by surface metrics such as EM or ROUGE~\citep{lin-2004-rouge}, even though these do not guarantee that an agent will make the same decision in the predicted state. Word2World \citep{li2025word} introduced the Real/WM/W2R protocol and Consistency Ratio (CR) for task-level replay consistency. We build on this protocol, introduce Pairwise Consistency Ratio ($\text{CR}_{pw}$) as our primary task-level criterion, and train WMs with a dense logged-action likelihood signal that asks whether the predicted state preserves support for the action taken in the real trajectory.

%% file: sections/03_formulation.tex
\section{Preliminaries}
\label{sec:formulation}

This section lays out the notation for describing the agent-environment interaction and presents a compact empirical study that quantifies the metric inversion phenomenon illustrated in Figure~\ref{fig:metric_inversion}.


\subsection{Agent--Environment Interaction}

\paragraph{Agent.}
The agent is defined as a policy $\pi$ that selects an action based on the interaction history at each step $t$:
\begin{equation}
  \pi : \{s_0,\, (a_i, s_i)_{i=1}^{t-1}\} \;\to\; a_t,
  \label{eq:agent}
\end{equation}
where $s_i$ is the observation returned after $a_i$, and $h_t=(s_0,a_1,s_1,\ldots,a_t)$.

\paragraph{World Model.}
As a surrogate for the real environment $\mathcal{E}$, the world model $\hat{\mathcal{E}}$ maps the history of textual observations and actions to the next observation and a task-completion signal:
\begin{equation}
  \hat{\mathcal{E}} : \{s_0,\, (a_i, \hat{s}_i)_{i=1}^{t-1},\, a_t\} \;\to\; (\hat{s}_{t},\, \hat{r}_{t}),
  \label{eq:wm}
\end{equation}
where $\hat{s}_{t}$ is the predicted next observation and $\hat{r}_{t}\in\{0,1\}$ is the task-completion signal.
In practice, $\hat{\mathcal{E}}$ is implemented as a single LLM trained on trajectory data from $\mathcal{E}$.

\subsection{Preliminary Observation: Decision-Critical Perturbations}
\label{sec:reward_validation}
Before introducing our method, we provide a simple empirical observation to illustrate the limitation of existing metrics.
To quantify the metric inversion phenomenon, we define \emph{Drop~Irrelevant} (DI) as a perturbation that removes non-critical content while preserving the next action and \emph{Drop~Target} (DT) as a perturbation that removes the target object and breaks the next action. A functionally meaningful metric should then rank DI above DT.
We compare BehR with action-consistency, LLM-judge, and text-similarity baselines; Appendix~\ref{sec:appendix_metric_definitions} defines each metric and gives its source. For readability in this diagnostic table, BehR is reported on a monotone $[0,1]$ scale where higher is better.

\begin{table}[t]
\centering
\footnotesize
\setlength{\tabcolsep}{3pt}
\resizebox{\columnwidth}{!}{%
\begin{tabular}{lcccc}
\toprule
{\bfseries Metric} & {\bfseries DI} & {\bfseries DT} & {\bfseries Rank?} & {\bfseries Dense?} \\
\midrule
{\bfseries BehR (Ours)} & 0.763 & 0.100 & \checkmark & \checkmark \\
ACS~\citep{guan2026computer} & 0.600 & 0.370 & \checkmark & $\times$ \\
GPT-4o Judge~\citep{zheng2023judging} & 0.466 & 0.824 & $\times$ & $\times$ \\
F1 Reward~\citep{wu2026rlvr} & 0.803 & 0.961 & $\times$ & $\times$ \\
BERTScore~\citep{zhang2019bertscore} & 0.859 & 0.974 & $\times$ & $\times$ \\
Exact Match & 0\% & 0\% & $\times$ & $\times$ \\
\bottomrule
\end{tabular}
}
\caption{\textbf{Decision-critical perturbation metrics.} Drop Irrelevant (DI) removes non-critical content, while Drop Target (DT) removes the target object and breaks the next action. A functionally meaningful metric should score DI above DT and provide a dense training signal.}
\label{tab:metric_ranking}
\end{table}

Table~\ref{tab:metric_ranking} confirms the metric-inversion pattern shown in Figure~\ref{fig:metric_inversion}. Text-based metrics and content-only LLM judges all rate the decision-critical DT perturbation as less severe because most tokens remain unchanged. BehR is the only metric that ranks the contrast correctly while serving as a usable dense training signal. The full seven-perturbation study for the non-judge automated metrics, together with definitions of all metrics used here, is deferred to Appendix~\ref{sec:appendix_perturbation}.

%% file: sections/04_method.tex
%
%
%
%
%
%

\section{Method}
\label{sec:method}
We propose behavior-consistency training for text-based world models as an alternative to \emph{state consistency}. Instead of only reconstructing the next observation, the world model should preserve the behavior induced by that observation: the predicted state should lead the agent to the same action distribution as the real state. Because black-box agents rarely expose full action distributions, we instantiate this property with Behavior Consistency Reward (BehR), a logged-action proxy optimized with reinforcement learning.


\subsection{Functional Consistency}
\label{sec:func_equiv}

We begin with the ideal objective: functional consistency.  



\begin{definition}[Functional Consistency]
A world model $\hat{\mathcal{E}}$ is functionally consistent with the real environment $\mathcal{E}$ with respect to an agent $\pi$ if the predicted state $\hat{s}_t$ induces the exact same action distribution as the real state $s_t$:
\begin{equation}
\pi(\cdot \mid h_t,\, \hat{s}_t) = \pi(\cdot \mid h_t,\, s_t), \quad \forall t.
\label{eq:fe}
\end{equation}
\end{definition}

Under this definition, two states are equivalent only when the agent's behavior is preserved, i.e., when the agent makes decisions according to the same distribution under either state. For black-box agents, however, we cannot access the full action distribution and thus cannot optimize Eq.~\ref{eq:fe} directly.


\begin{figure*}[t]
\centering
\includegraphics[width=\textwidth]{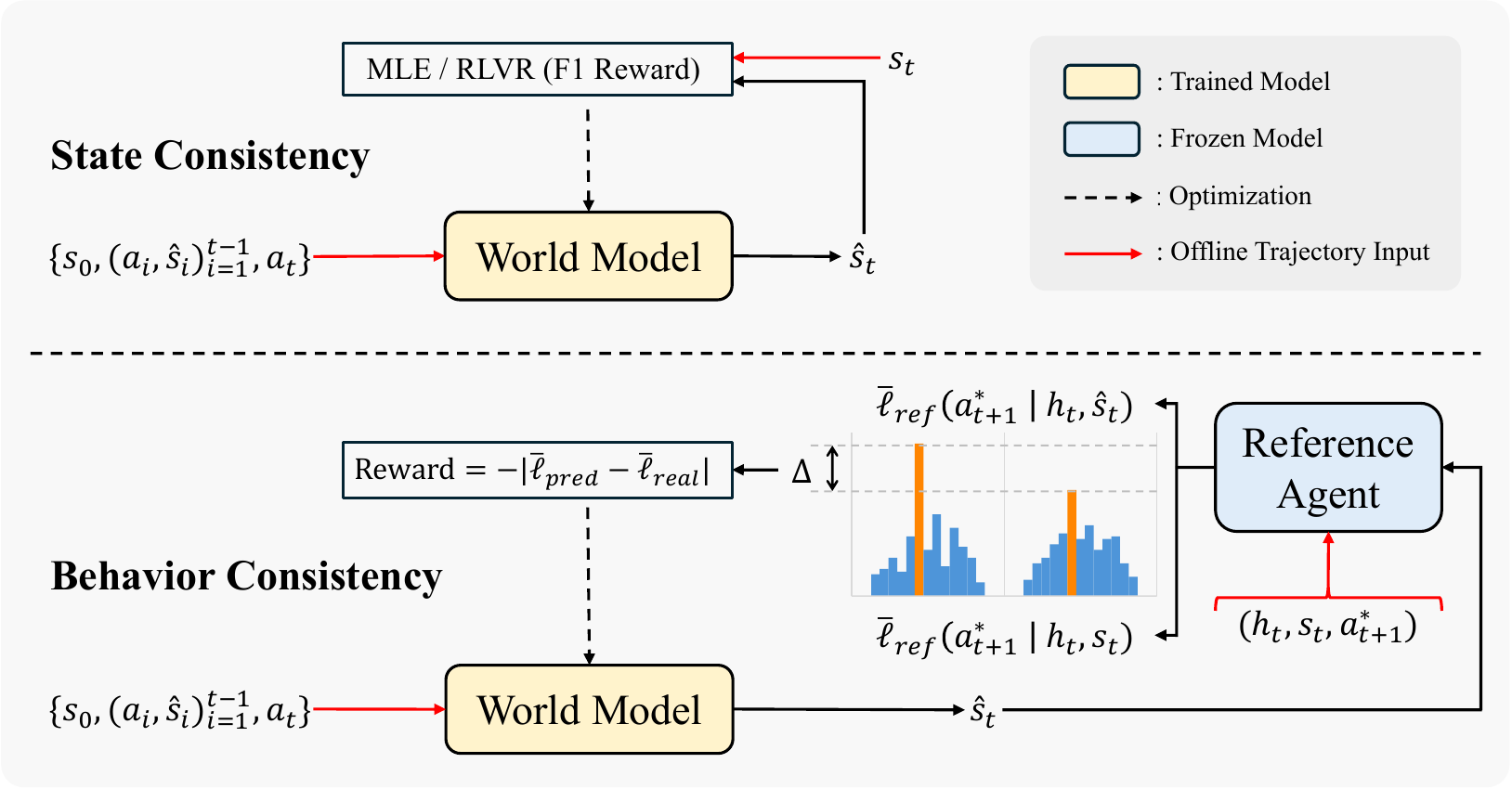}
\caption{\textbf{State consistency vs. behavior consistency training.} State-consistency objectives train the world model to match $s_t$, the real state after $a_t$, whereas BehR preserves the logged-action likelihood induced by that state. A frozen Reference Agent compares $a^*_{t+1}$ under $\hat{s}_t$ and $s_t$; their log-likelihood gap provides the reward.}
\label{fig:pipeline}
\end{figure*}

\subsection{Behavior Consistency Reward (BehR)}
\label{sec:behr}


We introduce Behavior Consistency Reward (BehR) as a tractable proxy for Eq.~\ref{eq:fe}. BehR does not recover the full action distribution; instead, it measures whether a frozen reference agent assigns similar likelihood to the logged next action under the predicted and real states.

\paragraph{Reference Agent.}
We use a \emph{frozen} external LLM $\pi_{\text{ref}}$ as a reference agent to estimate action likelihoods. Given history $h_t$, state $s_t$, and next action $a_{t+1}$, it outputs the mean per-token log-probability:
\begin{equation}
\bar{\ell}_{\text{ref}} \;\triangleq\; \frac{1}{|a_{t+1}|}\,\log \pi_{\text{ref}}(a_{t+1} \mid h_t,\, s_t),
\label{eq:reference}
\end{equation}
where $|a_{t+1}|$ is the token count of $a_{t+1}$. To avoid optimizing directly against task-level evaluation outcomes, $\pi_{\text{ref}}$ is kept fixed throughout training and used only to compute logged-action likelihoods. In our main experiments, Qwen3-8B serves as the Reference Agent; because some evaluated agents belong to the same family, we separately analyze reference-agent sensitivity in Table~\ref{tab:cross_ref_main} and Appendix~\ref{sec:appendix_judge_confound}.
BehR does not train the world model to imitate the reference policy: it depends on the relative likelihood difference of a logged action between predicted and real states, focusing the learning signal on how state changes affect support for a demonstrated decision rather than on absolute action preferences. This also avoids optimizing directly against the evaluated agents; in our experiments, the resulting WMs transfer across several downstream agents, although the strength of transfer still depends on the Reference Agent (Table~\ref{tab:cross_ref_main}).

\paragraph{Offline Training Data.}
Rewards are computed from offline trajectories without online interaction. Each tuple $(h_t, s_t, a^*_{t+1})$ contains the action-ending history, real state, and recorded next action. The real state serves as the behavioral anchor, while the logged next action probes whether a predicted state preserves behavior.

\paragraph{Reward Definition.}
Given a WM prediction $\hat{s}_t$, the frozen reference agent scores both the predicted and real states:
\begin{align}
\bar{\ell}_{\text{pred}} &= \bar{\ell}_{\text{ref}}(a^*_{t+1} \mid h_t,\, \hat{s}_t), \label{eq:lpred} \\
\bar{\ell}_{\text{real}} &= \bar{\ell}_{\text{ref}}(a^*_{t+1} \mid h_t,\, s_t). \label{eq:lreal}
\end{align}
We define BehR as
\begin{equation}
R_{\text{beh}} \;=\; -\bigl|\,\bar{\ell}_{\text{pred}} - \bar{\ell}_{\text{real}}\,\bigr|.
\label{eq:behr}
\end{equation}
$R_{\text{beh}} \le 0$ is maximized at zero when the two states induce identical likelihoods.
In implementation, we pass the monotone bounded score $\exp(R_{\text{beh}}) \in (0,1]$ to GRPO to control the numerical reward range; the underlying objective remains minimizing the absolute likelihood gap.
This proxy focuses on preserving decision-relevant actions observed in data, rather than matching the full decision boundary of the agent. Using the logged action also avoids scoring many candidate actions with the Reference Agent and reduces the risk of training toward its full action-ranking preferences.
This design makes the objective tractable under black-box agents, focuses the training signal on decision-critical behavior, and provides a stable and dense reward for optimization.

The absolute likelihood gap is symmetric: it penalizes states that remove evidence for the logged action and states that make the action artificially easier. This matters for simulator calibration: a WM that omits constraints or makes success cues overly salient may increase task success inside the WM while harming W2R transfer. BehR therefore encourages the WM to preserve the behavioral difficulty of the real transition, rather than merely producing a plausible or optimistic continuation.
\subsection{Behavior Consistency Training for World Models}

We optimize the world model using reinforcement learning with BehR as the training signal. 
Specifically, we use Group Relative Policy Optimization (GRPO)~\citep{shao2024deepseekmath}, which eliminates the critic network and stabilizes training by normalizing rewards within each prompt group.

For every prompt $h_t$, the WM generates $n$ candidate next states. BehR is computed for each candidate and normalized within the group. This group-relative normalization is useful because raw log-probability gaps are not directly comparable across prompts: some states naturally make the logged action very predictable, while others involve ambiguous or long actions with lower likelihoods. GRPO exploits this within-prompt comparison, rewarding candidates that better preserve the real-state action likelihood without requiring a separate value model.

Figure~\ref{fig:pipeline} summarizes the full training pipeline. The hyperparameters are given in Appendix~\ref{sec:appendix_impl}. 

%% file: sections/05_experiments.tex
\section{Experiments}
\label{sec:experiments}

\subsection{Experimental Setup}
\label{sec:setup}

\paragraph{Environments.} We evaluate world models on WebShop~\citep{yao2022webshop} (e-commerce) and TextWorld~\citep{cote2018textworld} (text-adventure).

\paragraph{Controlled comparison.} We use Word2World (W2W; Qwen2.5-7B) \citep{li2025word, qwen2.5}, a strong SFT baseline, as the base model, and initialize each BehR model from this checkpoint before post-training it with GRPO on a difficulty-filtered subset of the original training corpus (\S\ref{sec:appendix_grpo_data}). This design ensures that BehR-WM and W2W-WM share the same base model and training corpus. We also include LLaMA3.1-8B \citep{grattafiori2024llama} rows in the single-step and task-level tables as a cross-architecture stress test. To verify BehR's contribution, we run matched GRPO ablations with alternative reward functions (\S\ref{sec:reward_ablation}).

\paragraph{Evaluation Agents and Reference Agent.} We use four evaluation agents spanning three families: Qwen3-8B, Qwen3-32B, GPT-4o, and GPT-5 (all $T{=}0$), and a frozen Qwen3-8B as the Reference Agent. This supports reference-agent-mediated transfer tests in Table~\ref{tab:main_cr}, but it does not by itself establish reference-family invariance; we return to that limitation in Appendix~\ref{sec:appendix_judge_confound} and the Limitations section.

\paragraph{Training data.}
In both domains, we construct GRPO training data as step-level tuples (history, action, returned state, next expert action) from the original W2W corpus, yielding 4,321 WebShop tuples and 6,000 TextWorld tuples.
Full data-construction details are deferred to Appendix~\ref{sec:appendix_grpo_data}.

\paragraph{World-model variants.}
Table~\ref{tab:main_cr} compares {\bfseries W2W-WM} (the Word2World SFT checkpoint), {\bfseries F1-WM} (a matched GRPO control with token-level F1 reward following RLVR-World~\citep{wu2026rlvr}), and {\bfseries BehR-WM}. An additional exact-match GRPO control (SurR-WM) appears in Appendix~\ref{sec:appendix_surr_control}.

\paragraph{Evaluation protocol.} We evaluate each model from two complementary perspectives: single-step EM on held-out transitions, and task-level evaluation under the Real, WM, and W2R pipelines described below. We use deterministic decoding ($T{=}0$) throughout the analysis, as stochastic sampling may artificially depress measured consistency.

All task-level results use 200 held-out initial tasks per domain. Appendix~\ref{sec:appendix_ci} reports suite-level sign tests and bootstrap analyses over the benchmark cells.
\paragraph{Evaluation metrics.}
\label{sec:cr}
Following \citet{li2025word}, we report the following pipelines and metrics:
\begin{itemize}[leftmargin=*, nosep, topsep=2pt, partopsep=0pt]
    \item \textbf{Real}: the agent acts in the real environment $\mathcal{E}$.
    \item \textbf{WM}: the agent acts in the world-model environment $\hat{\mathcal{E}}$.
    \item \textbf{W2R}: the action sequence generated in the WM is replayed in the real environment $\mathcal{E}$.
    \item \textbf{Consistency Ratio (CR)}: the aggregate ratio $\text{CR} = \text{SR}_{\text{W2R}} / \text{SR}_{\text{Real}}$, where $\text{SR}$ denotes task success rate.
    \item \textbf{Pairwise Consistency Ratio (CR$_{\text{pw}}$)}: the fraction of individually Real-successful tasks that remain successful under W2R replay.
\end{itemize}
We treat CR$_{\text{pw}}$ as the primary metric and use aggregate CR as a complementary calibration diagnostic.

\subsection{Single-Step Accuracy}
\label{sec:single_step}

Before the task-level results, we verify that BehR-based optimization does not trade away local prediction quality. BehR preserves or improves held-out exact match in three of four domain--backbone settings (Table~\ref{tab:single_step}); nevertheless, high single-step EM does not guarantee task-level functional consistency \citep{li2025word}.

\begin{table}[t]
\centering
\small
\setlength{\tabcolsep}{6pt}
\begin{tabular*}{\columnwidth}{l@{\extracolsep{\fill}}lc}
\csname toprule\endcsname
{\bfseries Base Model} & {\bfseries World Model} & {\bfseries Exact Match (\%)} \\
\midrule
\multicolumn{3}{l}{\textit{WebShop (N=2,126 test samples)}} \\
\midrule
\multirow{2}{*}{Qwen2.5-7B}  & W2W-WM           & 79.05\% \\
                            & BehR-WM          & 79.19\% \\
\addlinespace
\multirow{2}{*}{LLaMA3.1-8B} & W2W-WM           & 77.37\% \\
                              & BehR-WM          & 75.97\% \\
\midrule
\multicolumn{3}{l}{\textit{TextWorld (N=1,993 test samples)}} \\
\midrule
\multirow{2}{*}{Qwen2.5-7B}  & W2W-WM           & 64.72\% \\
                            & BehR-WM          & 73.11\% \\
\addlinespace
\multirow{2}{*}{LLaMA3.1-8B} & W2W-WM           & 69.54\% \\
                              & BehR-WM          & 72.70\% \\
\bottomrule
\end{tabular*}
\caption{{\bfseries Single-step exact match.} Exact-match accuracy on held-out transitions for W2W-WM and BehR-WM across domains and backbones.}
\label{tab:single_step}
\end{table}

\subsection{Task-Level Functional Consistency}
\label{sec:main_results}

We now turn to the main results: task-level functional consistency.

\begin{table*}[!t]
\centering
\begingroup
\scriptsize
\setlength{\tabcolsep}{4pt}
\renewcommand{\arraystretch}{0.82}
\setlength{\aboverulesep}{0.25ex}
\setlength{\belowrulesep}{0.25ex}
\resizebox{0.98\textwidth}{!}{%
\begin{tabular}{@{\extracolsep{\fill}}ll ccccc ccccc}
\toprule
& & \multicolumn{5}{c}{\textbf{TextWorld}} & \multicolumn{5}{c}{\textbf{WebShop}} \\
\cmidrule(lr){3-7} \cmidrule(lr){8-12}
\textbf{Agent} & \textbf{World Model} & \textbf{Real} & \textbf{WM} & \textbf{W2R} & \textbf{CR} & \textbf{CR$_{\text{pw}}$} & \textbf{Real} & \textbf{WM} & \textbf{W2R} & \textbf{CR} & \textbf{CR$_{\text{pw}}$} \\
\midrule
\multicolumn{12}{l}{\textit{Qwen2.5-7B Base World Model}} \\
\midrule
\multirow{3}{*}{Qwen3-8B}
    & W2W-WM           & 87.0\% & 100.0\% & 64.5\% & 0.741          & 0.678          & 14.5\% & 16.5\% & 12.0\% & 0.828          & 0.345 \\
    & F1-WM            & 87.0\% &  97.5\% & 67.5\% & 0.776          & 0.690          & 14.5\% & 14.0\% &  9.5\% & 0.655          & 0.310 \\
    & \textbf{BehR-WM} & 87.0\% &  97.0\% & 67.5\% & 0.776 & \textbf{0.730} & 14.5\% & 17.0\% & 13.5\% & 0.931 & \textbf{0.483} \\
\midrule
\multirow{3}{*}{Qwen3-32B}
    & W2W-WM           & 97.0\% & 100.0\% & 49.0\% & 0.505          & 0.500          & 16.5\% & 19.0\% & 14.5\% & 0.879          & 0.455 \\
    & F1-WM            & 97.0\% &  99.5\% & 51.0\% & 0.526          & 0.521          & 16.5\% & 14.0\% & 12.0\% & 0.727          & 0.424 \\
    & \textbf{BehR-WM} & 97.0\% &  99.5\% & 52.0\% & 0.536 & \textbf{0.536} & 16.5\% & 15.5\% & 15.0\% & 0.909 & \textbf{0.485} \\
\midrule
\multirow{3}{*}{GPT-4o}
    & W2W-WM           & 99.5\% & 100.0\% & 99.0\% & 0.995          & \textbf{0.990} & 19.0\% & 19.0\% & 17.5\% & 0.921          & 0.760 \\
    & F1-WM            & 99.5\% &  93.0\% & 92.5\% & 0.930          & 0.925          & 19.0\% & 22.5\% & 18.5\% & 0.974 & 0.763 \\
    & \textbf{BehR-WM} & 99.5\% &  99.0\% & 99.0\% & 0.995 & \textbf{0.990} & 19.0\% & 21.0\% & 20.0\% & 1.053          & \textbf{0.840} \\
\midrule
\multirow{3}{*}{GPT-5}
    & W2W-WM           & 100.0\% & 100.0\% & 100.0\% & 1.000 & \textbf{1.000} & 39.0\% & 39.0\% & 35.5\% & 0.910          & 0.730 \\
    & F1-WM            & 100.0\% & 100.0\% & 100.0\% & 1.000 & \textbf{1.000} & 39.0\% & 43.5\% & 41.5\% & 1.064          & \textbf{0.740} \\
    & \textbf{BehR-WM} & 100.0\% &  99.5\% & 100.0\% & 1.000 & \textbf{1.000} & 39.0\% & 43.5\% & 37.5\% & 0.962 & \textbf{0.740} \\
\midrule
\multicolumn{12}{l}{\textit{LLaMA3.1-8B Base World Model}} \\
\midrule
\multirow{3}{*}{Qwen3-8B}
    & W2W-WM           & 87.0\% & 82.5\% & 55.5\% & 0.638          & 0.563          & 14.5\% & 27.5\% & 12.0\% & 0.828 & \textbf{0.345} \\
    & F1-WM            & 87.0\% & 89.5\% & 59.0\% & 0.678 & 0.563          & 14.5\% & 15.0\% & 11.5\% & 0.793          & 0.276 \\
    & \textbf{BehR-WM} & 87.0\% & 90.0\% & 58.5\% & 0.672          & \textbf{0.621} & 14.5\% & 12.5\% & 10.5\% & 0.724          & \textbf{0.345} \\
\midrule
\multirow{3}{*}{Qwen3-32B}
    & W2W-WM           & 97.0\% & 92.0\% & 63.0\% & 0.649          & 0.634          & 16.5\% & 13.5\% & 13.0\% & 0.788          & 0.485 \\
    & F1-WM            & 97.0\% & 98.0\% & 66.5\% & 0.686          & 0.675          & 16.5\% & 17.5\% & 12.0\% & 0.727          & 0.364 \\
    & \textbf{BehR-WM} & 97.0\% & 99.0\% & 69.0\% & 0.711 & \textbf{0.706} & 16.5\% & 17.5\% & 14.0\% & 0.848 & \textbf{0.515} \\
\midrule
\multirow{3}{*}{GPT-4o}
    & W2W-WM           & 99.5\% & 99.5\% & 94.5\% & 0.950          & 0.945          & 19.0\% & 19.0\% & 17.5\% & 0.921 & 0.710 \\
    & F1-WM            & 99.5\% & 95.5\% & 94.5\% & 0.950          & 0.950          & 19.0\% & 23.5\% & 22.0\% & 1.158          & 0.816 \\
    & \textbf{BehR-WM} & 99.5\% & 99.0\% & 99.0\% & 0.995 & \textbf{0.990} & 19.0\% & 21.5\% & 21.0\% & 1.105          & \textbf{0.890} \\
\midrule
\multirow{3}{*}{GPT-5}
    & W2W-WM           & 100.0\% &  99.0\% &  93.5\% & 0.935          & 0.935          & 39.0\% & 36.0\% & 34.5\% & 0.885          & 0.690 \\
    & F1-WM            & 100.0\% & 100.0\% &  99.5\% & 0.995 & \textbf{0.990} & 39.0\% & 44.0\% & 41.0\% & 1.051          & 0.690 \\
    & \textbf{BehR-WM} & 100.0\% & 100.0\% &  99.0\% & 0.990          & \textbf{0.990}          & 39.0\% & 41.5\% & 36.5\% & 0.936 & \textbf{0.720} \\
\bottomrule
\end{tabular}
}
\endgroup
\caption{\textbf{Task-level functional consistency.} Real/WM/W2R are success rates in the real environment, simulated environment, and real-environment replay of WM actions. CR = W2R/Real; CR$_{\text{pw}}$ is the fraction of Real-successful tasks preserved under W2R and is our primary metric. Bold marks the best CR$_{\text{pw}}$ value within each agent/backbone/domain group. F1-WM is a matched GRPO control; the additional SurR-WM control is reported in Appendix~\ref{sec:appendix_surr_control}.}
\label{tab:main_cr}
\end{table*}

Table~\ref{tab:main_cr} shows the clearest gains in the primary Qwen-base WebShop setting: BehR raises CR$_{\text{pw}}$ from $0.345 \to 0.483$ for Qwen3-8B, $0.455 \to 0.485$ for Qwen3-32B, $0.760 \to 0.840$ for GPT-4o, and $0.730 \to 0.740$ for GPT-5. The strongest improvements are therefore concentrated in WebShop and weak-to-mid-strength evaluation regimes rather than being uniformly large everywhere; small deltas such as GPT-5 on WebShop should be read as part of the aggregate pattern rather than as standalone evidence. Because CR is an aggregate ratio, it can exceed 1 when WM-generated action sequences succeed on more tasks under real-environment replay than the original Real run. We interpret such cases as over-optimistic aggregate calibration rather than as evidence that the WM improves the real environment, and therefore treat CR$_{\text{pw}}$ as the primary consistency metric.

The F1-WM rows provide a matched GRPO control using a surface-overlap reward. On WebShop this control is mixed: F1-WM reduces CR$_{\text{pw}}$ for Qwen3-8B ($0.310$ vs. $0.345$) and Qwen3-32B ($0.424$ vs. $0.455$), while BehR remains above or tied with both W2W and F1 across all 16 cells. This suggests that the reward target, not GRPO alone, drives the task-level gains. Appendix~\ref{sec:appendix_surr_control} reports the additional exact-match SurR-WM control; in the full four-method comparison, BehR is sole or tied top in all 16 cells.

TextWorld is harder to interpret because several baseline rows are near ceiling, so BehR often preserves high consistency rather than creating large new gains. Across all 16 W2W-vs.-BehR configurations, BehR improves CR$_{\text{pw}}$ in 13 rows and ties in 3; Appendix~\ref{sec:appendix_ci} reports suite-level statistical analyses for this aggregate pattern. Because WebShop WM rollouts can terminate late or reach the step cap, Appendix~\ref{sec:appendix_terminated_only} also reports a terminated-only check; the BehR ranking is preserved, suggesting that the main ordering is not driven by timeout handling.

To test whether gains are tied to Qwen3-8B as Reference Agent, Table~\ref{tab:cross_ref_main} retrains Qwen2.5-7B WebShop BehR-WMs with Ministral-3-8B and LLaMA3.1-8B while holding the base WM, data, and evaluation protocol fixed. The alternative-reference BehR-WMs match or exceed W2W-WM in all 8 configurations.

\begin{table}[t]
\centering
\small
\setlength{\tabcolsep}{3pt}
\resizebox{\columnwidth}{!}{%
\begin{tabular}{lccccc}
\csname toprule\endcsname
{\bfseries Eval Agent} & \textbf{W2W} & \textbf{F1} & \textbf{Qwen3-8B} & \textbf{Ministral-3-8B} & \textbf{LLaMA3.1-8B} \\
\midrule
Qwen3-8B  & 0.345 & 0.310 & \textbf{0.483} & 0.345 & 0.379 \\
Qwen3-32B & 0.455 & 0.424 & 0.485 & \textbf{0.576} & 0.455 \\
GPT-4o    & 0.760 & 0.763 & \textbf{0.840} & 0.763 & 0.763 \\
GPT-5     & 0.730 & 0.740 & 0.740 & \textbf{0.782} & 0.730 \\
\bottomrule
\end{tabular}
}
\caption{\textbf{Reference-agent ablation on WebShop.} CR$_{\text{pw}}$ for W2W/F1 controls and BehR-WMs trained with different Reference Agents; higher is better.}
\label{tab:cross_ref_main}
\end{table}

These results should not be read as full Reference-Agent invariance: the best reference varies across evaluation agents, and weaker references can produce smaller gains. The important point for our main claim is narrower but useful: the BehR signal is not unique to the original Qwen3-8B Reference Agent, and alternative open-weight references still avoid degrading below W2W in this WebShop ablation.

\subsection{Why the Optimization Target Matters}
\label{sec:reward_ablation}


To isolate the optimization target, we keep GRPO fixed and vary only the reward. As shown in Table~\ref{tab:main_cr}, F1-WM can reduce CR$_{\text{pw}}$ for weaker WebShop agents, while BehR-WM improves or matches W2W in the corresponding configurations. The additional exact-match SurR-WM control in Appendix~\ref{sec:appendix_surr_control} follows the same overall pattern. Thus surface-level or token-level rewards provide only partial signal, whereas BehR captures decision-critical information that surface matching misses.

%% file: sections/06_applications.tex
%
%
%
%
%
%
%
%

\section{Downstream Applications}
\label{sec:downstream}

We briefly test whether improved functional consistency makes WMs more useful as offline evaluators and planning simulators. These applications are where the distinction between surface fidelity and behavioral fidelity becomes operational: a WM that produces plausible text but changes agent outcomes can yield misleading leaderboards or unreliable planning rollouts.

\subsection{Application I: Calibrated Surrogate Evaluation}
\label{sec:surrogate_eval}

A useful surrogate should preserve ranking and avoid false positives. The ranking criterion asks whether the WM can recover the same relative ordering of agents, while calibration asks whether task-level successes and failures mean the same thing in the WM and in the real environment. Both WMs preserve aggregate leaderboard order well ($\rho=0.946/0.958$ on WebShop and $\rho=0.846/0.898$ on TextWorld for W2W/BehR; full leaderboard in Appendix~\ref{sec:appendix_trajectory}), so Table~\ref{tab:task_agreement} focuses on task-level agreement. Here FP denotes WM-only successes and Agree is overall agreement.

\begin{table}[t]
\centering
\small
\setlength{\tabcolsep}{3pt}
\begin{tabular*}{\columnwidth}{l@{\extracolsep{\fill}}lcccc}
\toprule
\textbf{Agent} & \textbf{WM} & \textbf{TP} & \textbf{TN} & \textbf{FP$\downarrow$} & \textbf{Agree} \\
\midrule
\multirow{2}{*}{\shortstack[l]{Qwen3\\0.6B}} & W2W-WM & 0 & 115 & 85 (42.5\%) & 57.5\% \\
 & BehR-WM & 0 & 181 & 19 \,(9.5\%) & \textbf{90.5\%} \\
\midrule
\multirow{2}{*}{\shortstack[l]{Qwen3\\1.7B}} & W2W-WM & 6 & 124 & 69 (34.5\%) & 65.0\% \\
 & BehR-WM & 5 & 159 & 34 (17.0\%) & \textbf{82.0\%} \\
\midrule
\multirow{2}{*}{GPT-5} & W2W-WM & 200 & 0 & 0 \,(0.0\%) & \textbf{100.0\%} \\
 & BehR-WM & 199 & 0 & 0 \,(0.0\%) & 99.5\% \\
\bottomrule
\end{tabular*}
\caption{\textbf{Task-level agreement on TextWorld.} Representative rows comparing WM and real-environment outcomes; FP denotes WM-only successes, and FN is omitted for compactness but can be recovered from the 200-task total.}
\label{tab:task_agreement}
\end{table}

The calibration gap is concentrated in weaker agents: W2W yields 42.5\%/34.5\% false positives for Qwen3-0.6B/1.7B, while BehR-WM reduces them to 9.5\%/17.0\% and raises agreement to 90.5\%/82.0\%. These false positives are tasks that appear solved inside the WM but fail after real-environment replay, causing weak agents to be systematically overestimated. This is the ``too easy simulator'' failure mode: a WM may smooth over constraints, omit blocking details, or produce completion-like trajectories that make weak agents look more capable than they are. BehR-WM reduces this failure mode without collapsing strong-agent performance. Additional breakdowns appear in Appendix~\ref{sec:appendix_trajectory}.

Calibrated WM evaluation is attractive because real-environment benchmarking can be slow and brittle, whereas WM evaluation reduces the loop to local batched inference. The benefit is meaningful only if WM successes transfer back to the real environment; task-level agreement and W2R replay therefore guard against fast but over-optimistic simulators.

\subsection{Application II: Lookahead Planning}
\label{sec:lookahead}

For planning, we use $K{=}5$ lookahead: the agent proposes candidate actions, the WM simulates next states, and the agent selects the most promising action. This turns the WM into a short-horizon decision aid rather than only an evaluator. The method costs $K{+}2$ model calls per step, so its usefulness depends on whether simulated states preserve action-relevant consequences.

\begin{table}[t]
\centering
\begingroup
\small
\setlength{\tabcolsep}{5pt}
\renewcommand{\arraystretch}{0.9}
\begin{tabular*}{\columnwidth}{l@{\extracolsep{\fill}}lcc}
\toprule
\textbf{Agent} & \textbf{Planning WM} & \textbf{SR} & \textbf{$\Delta$ vs.\ Base} \\
\midrule
\multicolumn{4}{l}{\textit{Qwen3-8B}} \\
& ReAct (base) & 14.5\% & --- \\
& W2W-WM & 24.0\% & +9.5pp \\
& BehR-WM & 25.5\% & +11.0pp \\
\midrule
\multicolumn{4}{l}{\textit{GPT-4o}} \\
 & ReAct (base) & 19.0\% & --- \\
 & W2W-WM & 26.0\% & +7.0pp \\
 & BehR-WM & 26.5\% & +7.5pp \\
\bottomrule
\end{tabular*}
\endgroup
\caption{\textbf{Lookahead planning on WebShop.} Success rates for ReAct and $K=5$ planning with W2W-WM or BehR-WM; $\Delta$ is the gain over ReAct.}
\label{tab:lookahead}
\end{table}

Table~\ref{tab:lookahead} shows that lookahead improves over ReAct and that BehR-WM further improves over W2W-WM in both shown pairs. The main gain comes from adding lookahead, while the incremental BehR-over-W2W gains are smaller and should be interpreted as evidence that better behavioral alignment can further improve the planning substrate. This pattern is consistent with our central claim: a WM is more useful for planning when its predicted futures preserve action-relevant consequences. Additional analysis appears in Appendix~\ref{sec:appendix_lookahead_main}.

%% file: sections/08_conclusion.tex
%
%

\section{Conclusion}
\label{sec:conclusion}

We argue that useful text-based world models should be functionally consistent with the real environment, not merely similar in surface form. This view yields both a task-level evaluation target (CR$_{\text{pw}}$) and a trainable step-level objective (BehR). Together, these components form a practical loop: identify decision-level failures with W2R/CR$_{\text{pw}}$, train against a behavior-sensitive reward, and test whether the resulting WM is better calibrated for evaluation and planning.

Empirically, BehR-based training improves task-level consistency in most of the 16 agent--domain--backbone configurations while largely preserving single-step prediction quality. The clearest gains appear in WebShop and weak-to-mid-strength evaluation regimes; near-ceiling TextWorld rows and the weakest LLaMA-based WebShop rows are more often tied or modest, while stronger WebShop agents can still show large gains. Downstream, BehR-WMs reduce weak-agent false positives ($42.5\% \to 9.5\%$ on TextWorld) and give the best point estimates in WebShop lookahead planning, suggesting that functional consistency is a practical criterion for simulator use.

\section*{Limitations}

The optimization-target comparison spans two domains, two backbones, four evaluation agents, and matched surface-reward controls, with suite-level statistics reported in Appendix~\ref{sec:appendix_ci}. Still, each GRPO configuration was trained once, so our analyses do not quantify run-to-run variance across independent training seeds. The GRPO runs also use one data-construction recipe and one core hyperparameter setting; broader decoding perturbations remain future work.

BehR is a logged-action proxy mediated by a frozen Reference Agent, so it does not fully capture the downstream agent's complete action distribution or all untested rollout regimes. Alternative-reference and cross-family analyses in Appendix~\ref{sec:appendix_judge_confound} mitigate reference-specific concerns, but broader reward families and a wider Reference-Agent sweep remain useful extensions.

\section*{Ethics Statement}

This work uses publicly available benchmarks and releases code and trained models. Surrogate evaluators can overestimate weak agents when the WM is too easy; such errors may distort capability comparisons across agent families or create misplaced confidence in simulator-only results. We therefore report calibration gaps and recommend W2R replay as the definitive validation step, especially before drawing deployment-facing conclusions from surrogate evaluation.

%% file: sections/09_appendix.tex
%
%

\appendix

\section{Environment and Dataset Details}
\label{sec:appendix_env}

We evaluate on two representative text-based interactive environments.
Task-level evaluation uses 200 held-out initial tasks per domain, drawn from the \textsc{AgentEval} benchmark suite of \citet{xi2024agentgym}, which provides standardized evaluation splits across diverse agent environments.
This count follows the AgentGym evaluation protocol: it refers to initial tasks rather than single-step samples.
Each initial task yields a multi-turn interaction trajectory, and in our setting the agent and world model typically interact for around ten steps on average before termination.
Table~\ref{tab:env_summary} summarizes their key characteristics.

\begin{table}[!h]
\centering
\small
\resizebox{\columnwidth}{!}{%
\begin{tabular}{@{}r@{\;:\;\;}p{0.60\columnwidth}@{}}
\toprule
\multicolumn{2}{@{}l}{\textbf{WebShop} \cite{yao2022webshop} (\textit{open-ended domain})} \\
\midrule
Domain    & E-commerce shopping \\
States    & Open-ended, partially observable \\
Actions   & \texttt{search[query]}, \texttt{click[element]} \\
\addlinespace[2pt]
Max steps & 50 \\
Success   & Purchase the correct item \\
Challenge & Diverse product pages, open catalog \\
\midrule
\multicolumn{2}{@{}l}{\textbf{TextWorld} \cite{cote2018textworld} (\textit{rule-governed domain})} \\
\midrule
Domain    & Text adventure game \\
States    & Structured, bounded \\
Actions   & NL commands (\texttt{go}, \texttt{take}, \ldots) \\
\addlinespace[2pt]
Max steps & 50 \\
Success   & Complete all subgoals \\
Challenge & Multi-room navigation, state tracking \\
\bottomrule
\end{tabular}
}
\caption{\textbf{Environment summary.} WebShop and TextWorld provide complementary open-ended and rule-governed tests for text-based world models.}
\label{tab:env_summary}
\end{table}

\subsection{WebShop}
\label{sec:appendix_webshop}

WebShop \cite{yao2022webshop} is a simulated e-commerce platform where an agent must find and purchase a product matching a natural-language instruction (e.g., ``Find me slim fit men's henleys with short sleeve, color: blue, size: medium, price lower than \$50'').
The environment provides search results pages listing multiple products (with ASINs, titles, and prices), item detail pages (with product descriptions, options, and reviews), and an action space consisting of \texttt{search[keywords]} and \texttt{click[element]} commands.
The agent must navigate through search, browse product pages, select correct options (color, size), and execute \texttt{click[buy now]} to complete the purchase.

\paragraph{World model challenges.}
The WM must generate realistic search result pages containing plausible product listings, maintain consistency of product attributes across navigation, and faithfully reproduce the purchase-completion signal.
In practice, WebShop presents the hardest WM challenge due to its open-ended state space: each search query can return different product combinations, and the WM must hallucinate coherent product catalogs that are internally consistent.

\subsection{TextWorld}
\label{sec:appendix_textworld}

TextWorld \cite{cote2018textworld} is a procedurally generated text-based interactive fiction framework.
Each game instance defines a set of interconnected rooms, objects with interactive affordances, and a sequence of subgoals (e.g., ``open the chest drawer, take the old key, unlock the wooden door, go east, take the milk from the refrigerator, place the milk on the stove'').
The agent receives natural-language observations describing the current room and available actions, and must issue single text commands per turn.

\paragraph{World model challenges.}
The WM must track object locations across multiple rooms, maintain inventory state, and correctly implement game-logic constraints (e.g., a locked door requires a specific key).
TextWorld's structured dynamics make it more amenable to world modeling than WebShop, but the multi-step dependency chains demand accurate state tracking over long horizons.

\subsection{GRPO Training Data Construction}
\label{sec:appendix_grpo_data}

Our BehR-based GRPO training data is derived from the original Word2World supervised world-model corpora through a step-level restructuring pipeline.
Figure~\ref{fig:data_pipeline} illustrates the overall process.

\begin{figure}[ht]
\centering
\small
\fbox{\parbox{0.92\columnwidth}{
\textbf{Step 1: Trajectory Collection.}
Expert agent (GPT-4o) interacts with real environments to collect multi-turn trajectories containing alternating actions and environment responses.
\\[4pt]
\textbf{Step 2: Step-Level Decomposition.}
Each trajectory is decomposed into individual transition tuples $(h_t, s_t, a^*_{t+1})$, where $h_t$ is the dialogue history ending with the current action $a_t$, $s_t$ is the returned real state, and $a^*_{t+1}$ is the logged next expert action.
\\[4pt]
\textbf{Step 3: Difficulty-Aware Filtering.}
For WebShop, we select hard samples (W2W baseline Token~F1 $<0.35$) and remove those with invalid expert actions, yielding 4,321 training samples.
For TextWorld, we allocate per-action-type budgets proportional to $\sqrt{\text{pool} \times \text{hardness}}$ across 632k step-level samples, yielding 6,000 training samples.
\\[4pt]
\textbf{Step 4: VeRL-Compatible Formatting.}
Each sample is formatted as a VeRL-compatible Parquet record with fields: \texttt{prompt} (WM message history ending at $a_t$), \texttt{reward\_model.ground\_truth} ($s_t$), and \texttt{extra\_info.expert\_action} ($a^*_{t+1}$).
}}
\caption{\textbf{GRPO training data construction pipeline.} The key contribution is the difficulty-aware filtering and step-level restructuring applied on top of Word2World's supervised corpora.}
\label{fig:data_pipeline}
\end{figure}

\paragraph{Data statistics.}
Table~\ref{tab:grpo_data_stats} provides detailed statistics of the resulting GRPO training datasets.

\begin{table}[!t]
\centering
\small
\begin{tabular*}{\columnwidth}{l@{\extracolsep{\fill}}cc}
\toprule
\textbf{Statistic} & \textbf{WebShop} & \textbf{TextWorld} \\
\midrule
Source trajectories & 70,790 & 58,805 \\
Train samples (step-level) & 4,321 & 6,000 \\
Test samples (step-level) & 2,126 & 1,993 \\
Avg.\ prompt length (tokens) & $\sim$2,500 & $\sim$1,800 \\
Avg.\ response length (tokens) & $\sim$350 & $\sim$80 \\
Action types & 2 & $\sim$8 \\
\bottomrule
\end{tabular*}
\caption{\textbf{GRPO data statistics.} Step-level train/test splits and sequence statistics after difficulty-aware filtering.}
\label{tab:grpo_data_stats}
\end{table}

\section{Implementation Details}
\label{sec:appendix_impl}

\subsection{BehR-Based GRPO Training Setup}

We train world models with the VeRL framework, invoking GRPO via its PPO trainer entry point.
Unless otherwise noted, the training runs discussed in the main text use the BehR reward described in Eq.~\ref{eq:behr}.
Representative main runs were conducted on 8$\times$A100 (80\,GB) GPUs and typically completed in approximately two days.

Table~\ref{tab:train_hparams} lists the core hyperparameters.
We use FSDP (Fully Sharded Data Parallelism) for model parallelism with bfloat16 precision, and vLLM as the rollout backend with tensor parallelism size 2.
The KL divergence penalty ($\beta{=}0.001$) prevents the WM from drifting too far from the W2W initialization.
We set the rollout temperature to 1.3, which is higher than typical LLM sampling.
The reason is that the base WM has already been trained via SFT on several hundred thousand trajectory samples, causing its output entropy to be extremely low under greedy or standard-temperature sampling.
At $T{=}1.0$, GRPO rollouts show near-zero entropy, meaning all $n{=}5$ candidates per prompt are nearly identical and provide no contrastive signal.
Raising the temperature to 1.3 restores entropy to a reasonable range ($\sim$0.1--1.0 nats/token), enabling GRPO to generate meaningfully diverse candidate states for reward comparison.

\begin{table}[ht]
\centering
\small
\setlength{\tabcolsep}{4pt}
\resizebox{\columnwidth}{!}{%
\begin{tabular}{p{0.34\columnwidth}p{0.50\columnwidth}}
\toprule
\textbf{Item} & \textbf{Value} \\
\midrule
Framework & VeRL (\texttt{verl}) \\
Trainer & \texttt{main\_ppo} \\
RL algorithm & GRPO \\
Learning rate & $5 \times 10^{-6}$ \\
Train batch size & 32 \\
PPO mini-batch size & 32 \\
Micro-batch / GPU & 2 (actor), 1 (rollout) \\
Max prompt length & 14,336 \\
Max response length & 1,024 \\
KL loss & on, coefficient $0.001$ \\
Gradient clipping & 1.0 \\
Rollout backend & vLLM \\
Tensor parallel size & 2 \\
Rollouts per prompt ($n$) & 5 \\
Rollout temperature & 1.3 \\
Top-$p$ & 1.0 \\
Model dtype & bfloat16 \\
FSDP opt. offload & on \\
Reward & BehR \\
Total epochs & 5 \\
Model save freq & every 20 steps \\
Hardware & 8$\times$A100 80\,GB \\
\bottomrule
\end{tabular}
}
\caption{\textbf{Training hyperparameters.} Core BehR-based GRPO configuration used for the main WebShop and TextWorld runs.}
\label{tab:train_hparams}
\end{table}

\subsection{Training Dynamics}
\label{sec:appendix_training_dynamics}

During BehR-based GRPO training, we observe the following dynamics:
\begin{itemize}
\item \textbf{WebShop}: The average BehR signal improves steadily over training, with the EM accuracy remaining stable ($\sim$79\%) throughout. This confirms that BehR optimization does not sacrifice single-step prediction quality.
\item \textbf{TextWorld}: The average BehR signal also improves, with EM accuracy notably improving from 64.7\% to 73.1\% (+8.4pp), suggesting that BehR-based training also regularizes toward more accurate predictions in structured domains.
\end{itemize}

\subsection{Model and Data Sources}

Table~\ref{tab:all_models} lists every model used in this paper together with its source.
The Qwen2.5-7B and LLaMA3.1-8B backbones each have a W2W variant from Word2World and a BehR variant introduced in this work. All four fine-tuned BehR WM variants have been publicly released and are summarized in Table~\ref{tab:model_ids}.
All Qwen3 \citep{yang2025qwen3} evaluation agents are used as-is without fine-tuning: Qwen3-8B and Qwen3-32B appear in the main CR comparison (Table~\ref{tab:main_cr}), while the full 0.6B--32B scale is used in surrogate evaluation (Table~\ref{tab:task_agreement} and Appendix~\ref{sec:appendix_webshop_analysis}).
Qwen3-8B additionally serves as the frozen BehR Reference Agent during the main training runs; Qwen3-32B is also the planner in lookahead experiments (\S\ref{sec:lookahead}).
GPT-4o and GPT-5 serve as proprietary evaluation agents; in W2R evaluation, their WM-generated action sequences are replayed in the real environment.

\begin{table*}[ht]
\centering
\small
\setlength{\tabcolsep}{12pt}
\resizebox{\textwidth}{!}{%
\begin{tabular}{@{}lll@{}}
\toprule
\textbf{Model} & \textbf{Role(s)} & \textbf{Source} \\
\midrule
\multicolumn{3}{@{}l}{\textit{World-Model Backbones (open-weight)}} \\
\midrule
Qwen2.5-7B  & Primary WM (W2W + BehR) & \texttt{Qwen/Qwen2.5-7B} \\
LLaMA3.1-8B & Cross-architecture WM (W2W + BehR) & \texttt{meta-llama/Llama-3.1-8B} \\
\midrule
\multicolumn{3}{@{}l}{\textit{Evaluation Agents (open-weight, Qwen3 series)}} \\
\midrule
Qwen3-0.6B  & Surrogate eval agent & \texttt{Qwen/Qwen3-0.6B} \\
Qwen3-1.7B  & Surrogate eval agent & \texttt{Qwen/Qwen3-1.7B} \\
Qwen3-4B    & Surrogate eval agent & \texttt{Qwen/Qwen3-4B} \\
Qwen3-8B    & Main eval agent + frozen BehR Reference Agent & \texttt{Qwen/Qwen3-8B} \\
Qwen3-14B   & Surrogate eval agent & \texttt{Qwen/Qwen3-14B} \\
Qwen3-32B   & Main eval agent + lookahead planner & \texttt{Qwen/Qwen3-32B} \\
\midrule
\multicolumn{3}{@{}l}{\textit{Alternative Reference Agents (open-weight)}} \\
\midrule
Ministral-3-8B & Alternative frozen Reference Agent & \texttt{mistralai/Ministral-3-8B-Instruct-2512} \\
LLaMA3.1-8B & Alternative frozen Reference Agent & \texttt{meta-llama/Llama-3.1-8B} \\
\midrule
\multicolumn{3}{@{}l}{\textit{API Models (proprietary)}} \\
\midrule
GPT-4o      & Main eval agent + W2R & OpenAI API (\texttt{gpt-4o-2024-11-20}) \\
GPT-5       & Main eval agent + W2R & OpenAI API (\texttt{gpt-5-2025-08-07}) \\
\bottomrule
\end{tabular}
}
\caption{\textbf{Models used in the experiments.} Roles and sources for world-model backbones, evaluation agents, alternative Reference Agents, and API models. Fine-tuned BehR WM release identifiers are summarized in Table~\ref{tab:model_ids}.}
\label{tab:all_models}
\end{table*}

\paragraph{Model availability and data provenance.}
All processed data used by our GRPO pipeline are derived from the original Word2World supervised world-model corpora; both domains contain step-level samples split into a training set for GRPO and a held-out test set for EM evaluation.
The Word2World checkpoints are public prior-work artifacts \citep{li2025word}, and their released identifiers are listed alongside the publicly released BehR checkpoints in Table~\ref{tab:model_ids}.
We release the code required to prepare the data from the upstream corpora, but do not separately redistribute the processed datasets.
Reference-agent model sources, including Ministral-3-8B and LLaMA3.1-8B, are listed in Table~\ref{tab:all_models}.

\begin{table*}[ht]
\centering
\small
\setlength{\tabcolsep}{3.5pt}
\resizebox{0.94\textwidth}{!}{%
\begin{tabular}{llll}
\toprule
\textbf{Domain} & \textbf{Base WM} & \textbf{W2W release} & \textbf{BehR release} \\
\midrule
WebShop & Qwen2.5-7B & \texttt{X1AOX1A/WorldModel-Webshop-Qwen2.5-7B} & \texttt{Ricardo-H/BehR-WorldModel-Webshop-Qwen2.5-7B} \\
WebShop & LLaMA3.1-8B & \texttt{X1AOX1A/WorldModel-Webshop-Llama3.1-8B} & \texttt{Ricardo-H/BehR-WorldModel-Webshop-Llama3.1-8B} \\
TextWorld & Qwen2.5-7B & \texttt{X1AOX1A/WorldModel-Textworld-Qwen2.5-7B} & \texttt{Ricardo-H/BehR-WorldModel-Textworld-Qwen2.5-7B} \\
TextWorld & LLaMA3.1-8B & \texttt{X1AOX1A/WorldModel-Textworld-Llama3.1-8B} & \texttt{Ricardo-H/BehR-WorldModel-Textworld-Llama3.1-8B} \\
\bottomrule
\end{tabular}
}
\caption{\textbf{World-model release summary.} Public Word2World checkpoint identifiers are shown; BehR world models are publicly released on Hugging Face at the identifiers above.}
\label{tab:model_ids}
\end{table*}

\section{BehR Reward Function Details}
\label{sec:appendix_behr}

The BehR reward is computed for each candidate world-model state $\hat{s}_t$ as follows:

\begin{enumerate}
\item \textbf{Reference-Agent prompt construction}: Build an agent-perspective prompt from the dialogue history $h_t$ and candidate state $\hat{s}_t$, ending with the logged expert action $a^*_{t+1}$ (see \S\ref{sec:appendix_judge_prompt}).

\item \textbf{Log-probability computation}: Query the frozen Reference Agent (Qwen3-8B) via the vLLM HTTP API to obtain per-token log-probabilities for $a^*_{t+1}$:
let $c_t = (h_t, \hat{s}_t)$ denote the agent context, and define
\begin{align*}
\bar{\ell}_{\text{ref}}(a^*_{t+1} \mid c_t)
  &= \frac{1}{|a^*_{t+1}|}\log \pi_{\text{ref}}(a^*_{t+1} \mid c_t)
\end{align*}

\item \textbf{Reward computation}: Compute the Reference-Agent likelihood difference $\Delta = \bar{\ell}_{\text{pred}} - \bar{\ell}_{\text{real}}$ and set
$$R_{\text{beh}} = -|\Delta|.$$
For numerical range control, the scalar passed to GRPO is the monotone bounded score $\exp(R_{\text{beh}}) \in (0,1]$.
\end{enumerate}

\paragraph{Real-state log-probability caching.}
A key efficiency optimization is real-state caching: under GRPO with $n{=}5$ rollouts per prompt, all 5 candidates share the same real state $s_t$ and thus the same $\bar{\ell}_{\text{real}}$.
For the representative batch configuration in Table~\ref{tab:train_hparams}, this reduces API calls from $32 \times 5 \times 2 = 320$ to $32 \times (5 + 1) = 192$ per batch, an approximately 40\% reduction.
The asymptotic saving approaches $1 - (n+1)/(2n)$ for group size $n$, so larger rollout groups would yield larger reductions.

\section{Additional Surface-Reward Control}
\label{sec:appendix_surr_control}

SurR-WM is an additional GRPO control trained with a surface exact-match reward, initialized from the same W2W-WM checkpoints and using the same step-level training data as F1-WM and BehR-WM. We keep it as a supplementary control because it was designed to test whether a simple surface reward is sufficient, whereas the main text focuses on the stronger prior-work-inspired F1 reward.

\begin{table*}[t]
\centering
\begingroup
\scriptsize
\setlength{\tabcolsep}{3.5pt}
\renewcommand{\arraystretch}{0.88}
\resizebox{0.92\textwidth}{!}{%
\begin{tabular}{lllcccc}
\csname toprule\endcsname
{\bfseries Domain} & {\bfseries Base WM} & {\bfseries Eval Agent} & {\bfseries W2W-WM} & {\bfseries F1-WM} & {\bfseries SurR-WM} & {\bfseries BehR-WM} \\
\midrule
\multirow{8}{*}{TextWorld}
& \multirow{4}{*}{Qwen2.5-7B}
& Qwen3-8B  & 0.678 & 0.690 & \textbf{0.730} & \textbf{0.730} \\
& & Qwen3-32B & 0.500 & 0.521 & 0.500 & \textbf{0.536} \\
& & GPT-4o    & \textbf{0.990} & 0.925 & 0.945 & \textbf{0.990} \\
& & GPT-5     & \textbf{1.000} & \textbf{1.000} & \textbf{1.000} & \textbf{1.000} \\
\cmidrule(lr){2-7}
& \multirow{4}{*}{LLaMA3.1-8B}
& Qwen3-8B  & 0.563 & 0.563 & 0.600 & \textbf{0.621} \\
& & Qwen3-32B & 0.634 & 0.675 & 0.696 & \textbf{0.706} \\
& & GPT-4o    & 0.945 & 0.950 & 0.945 & \textbf{0.990} \\
& & GPT-5     & 0.935 & \textbf{0.990} & \textbf{0.990} & \textbf{0.990} \\
\midrule
\multirow{8}{*}{WebShop}
& \multirow{4}{*}{Qwen2.5-7B}
& Qwen3-8B  & 0.345 & 0.310 & 0.345 & \textbf{0.483} \\
& & Qwen3-32B & 0.455 & 0.424 & 0.394 & \textbf{0.485} \\
& & GPT-4o    & 0.760 & 0.763 & 0.816 & \textbf{0.840} \\
& & GPT-5     & 0.730 & \textbf{0.740} & 0.731 & \textbf{0.740} \\
\cmidrule(lr){2-7}
& \multirow{4}{*}{LLaMA3.1-8B}
& Qwen3-8B  & \textbf{0.345} & 0.276 & 0.310 & \textbf{0.345} \\
& & Qwen3-32B & 0.485 & 0.364 & 0.485 & \textbf{0.515} \\
& & GPT-4o    & 0.710 & 0.816 & 0.684 & \textbf{0.890} \\
& & GPT-5     & 0.690 & 0.690 & 0.706 & \textbf{0.720} \\
\bottomrule
\end{tabular}
}
\endgroup
\caption{\textbf{Additional surface-reward GRPO control.} CR$_{\text{pw}}$ for the exact-match reward control (SurR-WM), shown alongside W2W-WM, F1-WM, and BehR-WM. SurR-WM provides an additional sanity check, while the main text emphasizes F1-WM as the stronger prior-work-inspired surface-overlap baseline.}
\label{tab:appendix_surr_control}
\end{table*}

\section{Terminated-Only WebShop Consistency}
\label{sec:appendix_terminated_only}

WebShop WM episodes often reach the 50-step cap, especially for weaker agents. To check whether CR$_{\text{pw}}$ rankings are driven by timeout handling, we also compute CR$_{\text{pw}}^{T}$ on the subset of WebShop tasks whose WM trajectory terminates normally before the cap. Table~\ref{tab:terminated_only} reports CR$_{\text{pw}}^{T}$ together with the WM-side timeout fraction.

\begin{table*}[t]
\centering
\begingroup
\scriptsize
\setlength{\tabcolsep}{3.5pt}
\renewcommand{\arraystretch}{0.88}
\resizebox{0.88\textwidth}{!}{%
\begin{tabular}{llcccc}
\csname toprule\endcsname
{\bfseries Base WM} & {\bfseries Agent} & {\bfseries W2W-WM} & {\bfseries F1-WM} & {\bfseries SurR-WM} & {\bfseries BehR-WM} \\
\midrule
Qwen2.5-7B  & Qwen3-8B  & 0.714 (84\%) & 0.692 (78\%) & 0.471 (56\%) & \textbf{0.824} (76\%) \\
Qwen2.5-7B  & Qwen3-32B & 0.882 (81\%) & 0.875 (84\%) & 0.600 (55\%) & \textbf{1.000} (75\%) \\
Qwen2.5-7B  & GPT-4o    & \textbf{1.000} (80\%) & \textbf{1.000} (77\%) & \textbf{1.000} (77\%) & \textbf{1.000} (78\%) \\
Qwen2.5-7B  & GPT-5     & 0.898 (62\%) & 0.903 (56\%) & 0.846 (55\%) & \textbf{0.906} (56\%) \\
\midrule
LLaMA3.1-8B & Qwen3-8B  & \textbf{0.769} (72\%) & \textbf{0.769} (76\%) & 0.562 (64\%) & \textbf{0.769} (78\%) \\
LLaMA3.1-8B & Qwen3-32B & 0.889 (79\%) & 0.750 (81\%) & 0.750 (54\%) & \textbf{1.000} (82\%) \\
LLaMA3.1-8B & GPT-4o    & \textbf{1.000} (81\%) & \textbf{1.000} (75\%) & 0.926 (78\%) & \textbf{1.000} (78\%) \\
LLaMA3.1-8B & GPT-5     & 0.845 (64\%) & 0.846 (57\%) & 0.859 (55\%) & \textbf{0.862} (59\%) \\
\bottomrule
\end{tabular}
}
\endgroup
\caption{\textbf{Terminated-only WebShop consistency.} CR$_{\text{pw}}^{T}$ is computed only on tasks whose WM trajectory terminates normally before the 50-step cap. Parentheses show the WM-side timeout fraction. Bold marks the sole or tied top value in each row.}
\label{tab:terminated_only}
\end{table*}

The terminated-only results preserve the main conclusion: BehR-WM is sole or tied top in all WebShop rows, so the CR$_{\text{pw}}$ ranking is not an artifact of counting timeout cases. We do not add a termination reward during training because doing so would optimize for shorter WM episodes rather than faithful transition dynamics; in WebShop, many timeouts reflect agent-side loops in long HTML contexts rather than a missing termination token.

\section{Step-Wise Behavioral Consistency}
\label{sec:appendix_stepwise}

CR$_{\text{pw}}$ is a task-level metric. As a complementary fine-grained check, we align WM-generated and real trajectories step by step under the evaluation agent's actual policy and measure action-pair agreement. This analysis includes suboptimal agent moves, so it tests whether the WM preserves local behavioral consequences beyond expert demonstrations.

\begin{table*}[t]
\centering
\begingroup
\scriptsize
\setlength{\tabcolsep}{3.5pt}
\renewcommand{\arraystretch}{0.88}
\resizebox{0.90\textwidth}{!}{%
\begin{tabular}{ll lccc}
\csname toprule\endcsname
{\bfseries Eval. Agent} & {\bfseries Domain} & {\bfseries Base WM} & {\bfseries W2W-WM} & {\bfseries F1-WM} & {\bfseries BehR-WM} \\
\midrule
Qwen3-8B & TextWorld & Qwen2.5-7B  & 31.2\% & 33.8\% & \textbf{34.7\%} \\
Qwen3-8B & WebShop   & Qwen2.5-7B  & 27.3\% & 28.9\% & \textbf{33.9\%} \\
Qwen3-8B & TextWorld & LLaMA3.1-8B & 33.3\% & 45.1\% & \textbf{45.3\%} \\
Qwen3-8B & WebShop   & LLaMA3.1-8B & 27.9\% & 35.5\% & \textbf{36.0\%} \\
\midrule
GPT-4o   & TextWorld & Qwen2.5-7B  & 80.2\% & 75.8\% & \textbf{83.1\%} \\
GPT-4o   & WebShop   & Qwen2.5-7B  & 68.6\% & 69.7\% & \textbf{71.1\%} \\
GPT-4o   & TextWorld & LLaMA3.1-8B & 81.3\% & 79.8\% & \textbf{82.4\%} \\
GPT-4o   & WebShop   & LLaMA3.1-8B & 56.2\% & 67.6\% & \textbf{71.9\%} \\
\bottomrule
\end{tabular}
}
\endgroup
\caption{\textbf{Step-wise behavioral consistency.} Action-pair agreement between aligned WM-generated and real trajectories. BehR-WM leads in all shown cells, providing fine-grained support beyond task-level CR$_{\text{pw}}$.}
\label{tab:stepwise_consistency}
\end{table*}

Step-wise pairs within a trajectory are correlated, so this table should be read as complementary diagnostic evidence rather than as an independent sample-size multiplier. Its value is showing that BehR's advantage appears at the action level as well as in task-level replay.

\section{Prompt Templates}
\label{sec:appendix_prompts}

\subsection{World Model Prompt Format}
\label{sec:appendix_wm_prompt}

The world model operates in a \emph{reversed} role convention compared to the agent:
the \texttt{user} role contains agent actions, while the \texttt{assistant} role contains
environment observations. This design allows the WM to generate the next environment
state conditioned on the action.

\begin{tcolorbox}[
title=WebShop World Model Prompt,
fonttitle=\bfseries,
colback=teal!5,
colbacktitle=teal!60,
colframe=teal!70!black,
coltitle=black,
boxrule=0.6pt,
arc=2mm,
breakable
]
\small
\begin{verbatim}
[System] WebShop [SEP] Instruction: [SEP]
  Find me <product description> with
  color: <color>, size: <size>,
  price lower than <price> [SEP] Search
  Your admissible actions: [...]

[User] search[<query>]

[Assistant] Instruction: [SEP] <task> [SEP]
  Back to Search [SEP] Page 1 (Total
  results: 50) [SEP] Next > [SEP]
  B09BNCFGLY [SEP] <product title 1>
  [SEP] $<price1> [SEP] ...
  Your admissible actions: [...]

[User] click[b09bncfgly]

[Assistant] <item detail page...>
\end{verbatim}
\end{tcolorbox}

\begin{tcolorbox}[
title=TextWorld World Model Prompt,
fonttitle=\bfseries,
colback=teal!5,
colbacktitle=teal!60,
colframe=teal!70!black,
coltitle=black,
boxrule=0.6pt,
arc=2mm,
breakable
]
\small
\begin{verbatim}
[System] <TextWorld ASCII banner>
  Welcome to TextWorld! <task description>
  -= Bedroom =-
  <room description>
  AVAILABLE ACTIONS: <action list>

[User] open chest drawer

[Assistant] You open the chest drawer,
  revealing an old key.
  -= Bedroom =- 1/3
  AVAILABLE ACTIONS: take old key, ...

[User] take old key from chest drawer

[Assistant] You pick up the old key.
  ...
\end{verbatim}
\end{tcolorbox}

\subsection{Agent System Prompts}
\label{sec:appendix_agent_prompt}

All agents follow the standard ReAct paradigm \cite{yao2023react} as implemented in AgentGym \cite{xi2024agentgym}, producing \texttt{Thought:} / \texttt{Action:} pairs at each turn.
The domain-specific system prompts below are loaded from \texttt{init\_contexts/} at evaluation time.
The BehR Reference Agent uses the \emph{same} system prompt (per domain) when constructing its agent-perspective prompt for log-probability computation (\S\ref{sec:appendix_judge_prompt}); it prepends the \texttt{Action:\textbackslash n} prefix so that log-probabilities are computed only over the action tokens, skipping the \texttt{Thought} portion.

\begin{tcolorbox}[
title=WebShop Agent Prompt,
fonttitle=\bfseries,
colback=blue!5,
colbacktitle=blue!60,
colframe=blue!70!black,
boxrule=0.6pt,
arc=2mm,
breakable
]
\ttfamily\small
You are web shopping.\\
I will give you instructions about what to do.\\
You have to follow the instructions.\\
Every round I will give you an observation and a list of
available actions, you have to respond an action based on
the state and instruction.\\
You can use search action if search is available.\\
You can click one of the buttons in clickables.\\
An action should be of the following structure:\\
search[keywords]\\
click[value]\\
If the action is not valid, perform nothing.\\
Keywords in search are up to you, but the value in click
must be a value in the list of available actions.\\
Remember that your keywords in search should be carefully
designed.\\
Your response should use the following format:\\[6pt]
Thought:\\
I think ...\\[6pt]
Action:\\
click[something]
\end{tcolorbox}

\begin{tcolorbox}[
title=TextWorld Agent Prompt,
fonttitle=\bfseries,
colback=blue!5,
colbacktitle=blue!60,
colframe=blue!70!black,
boxrule=0.6pt,
arc=2mm,
breakable
]
\ttfamily\small
You are playing a text-based interactive fiction game
(TextWorld).\\
You will receive observations describing the current state.\\
When available, a list of admissible actions may be provided.\\
Always output strictly in the following format:\\[6pt]
``Thought:\\
<your reasoning>\\[6pt]
Action:\\
<the single action to take>''\\[6pt]
Guidelines:\\
- Prefer actions from admissible commands when provided.\\
- If no list is provided, issue a valid single command
  (e.g., ``look'', ``inventory'', ``open door'', ``go north'',
  ``take key'').\\
- Avoid invalid or multiple actions in one step.
\end{tcolorbox}

\subsection{Reference-Agent Prompt Construction}
\label{sec:appendix_judge_prompt}

The Reference Agent reuses the same agent system prompt (\S\ref{sec:appendix_agent_prompt}) and the same ReAct format used during agent evaluation.
The WM-generated candidate state $\hat{s}_t$ is placed as the final user turn, and an \texttt{Action:\textbackslash n} prefix is appended as the assistant turn so that log-probabilities are computed only over the expert action tokens $a^*_{t+1}$, skipping the \texttt{Thought} portion.
We disable Qwen3's thinking mode (\texttt{enable\_thinking=False}) to prevent \texttt{<think>} tokens from interfering with log-probability computation.

\subsection{Lookahead Planning Prompts}
\label{sec:appendix_lookahead_prompt}

The local lookahead planner (\S\ref{sec:lookahead}) uses a two-stage prompt protocol at each step: \emph{candidate proposal} selects $K$ promising actions from the admissible set, and \emph{best-action selection} chooses the final action after observing WM-predicted futures.
Both prompts are domain-generic; the only domain-specific element is the agent system prompt inherited from the interaction context.

\paragraph{Step-by-step procedure.}
At each decision step $t$, the planner executes:
\begin{enumerate}[nosep,leftmargin=*]
\item \textbf{Candidate proposal.} The planner LLM receives the current observation and the full list of admissible actions, and outputs the top-$K$ actions ranked by estimated promise (Stage~1 prompt below). This costs one LLM call.
\item \textbf{WM rollout.} For each of the $K$ candidate actions, the world model generates a predicted next state $\hat{s}_{t+1}^{(k)}$, $k=1,\dots,K$. This costs $K$ WM calls (batched).
\item \textbf{Best-action selection.} The planner LLM receives all $K$ (action, predicted state) pairs and selects the action whose predicted outcome best advances the task goal (Stage~2 prompt below). This costs one LLM call.
\item \textbf{Execution.} The selected action is sent to the real environment (or WM); the returned observation becomes the context for step $t{+}1$.
\end{enumerate}
\noindent Total inference cost per step: $K{+}2$ LLM calls ($K$ WM + $2$ planner).
With $K{=}5$, this is $7{\times}$ the cost of standard ReAct, which motivates keeping $K$ small.

\paragraph{Stage 1: Candidate proposal.}
\begin{tcolorbox}[
title=Lookahead Candidate Proposal,
fonttitle=\bfseries,
colback=orange!5,
colbacktitle=orange!60,
colframe=orange!70!black,
boxrule=0.6pt,
arc=2mm,
breakable
]
\ttfamily\small
You are currently in this state:\\
\{observation\}\\[6pt]
All admissible actions:\\
\{numbered list\}\\[6pt]
Your task is described in the instruction above.\\
From the admissible actions, select the K actions that are
MOST LIKELY to help you complete the task successfully.\\[6pt]
Output EXACTLY K actions, one per line, in the format:\\
1. action\_here\\
2. action\_here\\
\ldots\\
Only output the numbered list, nothing else.
\end{tcolorbox}

\paragraph{Stage 2: Best-action selection.}
\begin{tcolorbox}[
title=Lookahead Best-Action Selection,
fonttitle=\bfseries,
colback=orange!5,
colbacktitle=orange!60,
colframe=orange!70!black,
boxrule=0.6pt,
arc=2mm,
breakable
]
\ttfamily\small
\textnormal{\textbf{[System]}} You are a decision-making assistant.
You will be given a current state and multiple action options
with their predicted outcomes. Select the BEST option.\\[6pt]
\textnormal{\textbf{[User]}} CURRENT STATE:\\
\{observation\}\\[6pt]
AVAILABLE OPTIONS (with predicted outcomes from a world
model):\\[4pt]
Option 1: \{action\}\\
\quad Predicted next state: \{preview\}\\
\ldots\\[6pt]
Select the option that best advances the task goal.\\
Reply with ONLY the option number (e.g., 1 or 3).
\end{tcolorbox}

\noindent For TextWorld, the selector outputs the exact action text instead of a number, since TextWorld actions are free-form commands rather than clickable elements.

\section{Case Study: Complete Trajectory Comparison}
\label{sec:appendix_cases}

Figure~\ref{fig:full_trajectory} presents a complete TextWorld trajectory (\texttt{textworld\_169}) comparing agent behavior under three conditions: the real environment, BehR-WM, and W2W-WM.
The agent is Qwen3-8B; the WM backbone is LLaMA3.1-8B, matching the cross-architecture setting reported in Table~\ref{tab:main_cr}.

\begin{figure*}[t]
\centering
\small
\fbox{\parbox{0.95\textwidth}{
\textbf{Task}: open trunk $\to$ take key $\to$ unlock door $\to$ open door $\to$ go east $\to$ open fridge $\to$ take lettuce $\to$ put lettuce on stove \quad (3 subgoals, 8 optimal steps)
\\[4pt]
\hrule
\vspace{4pt}
{\bfseries Steps 1--7: Shared prefix}: all three conditions produce identical agent actions and functionally equivalent WM responses.
\\[2pt]
\centering
\begin{tabular}{clp{0.60\textwidth}}
\toprule
\textbf{\#} & \textbf{Agent Action} & \textbf{WM / Env Response (abbreviated)} \\
\midrule
1 & \texttt{open antique trunk} & ``You open the antique trunk, revealing an old key.'' \\
2 & \texttt{take old key} & ``You take the old key from the antique trunk.'' \\
3 & \texttt{unlock wooden door} & ``(with the old key) You unlock wooden door.'' \\
4 & \texttt{open wooden door} & ``You open wooden door.'' \\
5 & \texttt{go east} & $-$= Kitchen $=$$-$\; \textit{(room description, refrigerator, stove, exits)} \quad Score +1 \\
6 & \texttt{open refrigerator} & ``You open the refrigerator, revealing \ldots\ and a lettuce.'' \\
7 & \texttt{take lettuce} & ``You take the lettuce from the refrigerator.'' \quad Score +1 \\
\bottomrule
\end{tabular}
\\[2pt]
\raggedright
{\itshape Minor surface variation}: BehR-WM fridge contains ``a bell pepper and a lettuce'' vs.\ real ``a half bag of chips and a lettuce'', with no behavioral impact since the agent only needs the lettuce.
\\[4pt]
\hrule
\vspace{4pt}
{\bfseries Step 8+: Divergence}: agent is in Kitchen with lettuce, score 2/3.
\\[4pt]
\colorbox{green!10}{\parbox{0.93\textwidth}{%
\textbf{Real Environment} \hfill \checkmark\; \textbf{Success (8 steps)}\\[1pt]
Agent $\to$ \texttt{put lettuce on stove} $\to$ ``You put the lettuce on the stove. \textit{*** The End ***}\; Score 3/3''
}}
\\[4pt]
\colorbox{green!10}{\parbox{0.93\textwidth}{%
\textbf{BehR-WM (LLaMA-8B)} \hfill \checkmark\; \textbf{Success (9 steps)}\\[1pt]
Agent $\to$ \texttt{rest lettuce on stove} $\to$ WM: ``That's not a verb I recognise.''
\quad {\color{gray}(correct error, same as real env)}\\
Agent $\to$ \texttt{put lettuce on stove} $\to$ ``You put the lettuce on the stove. \textit{*** The End ***}\; Score 3/3''
}}
\\[4pt]
\colorbox{red!10}{\parbox{0.93\textwidth}{%
{\bfseries W2W-WM (LLaMA-8B)} \hfill $\times$\; {\bfseries Fail (timeout at 50 steps)}\\[1pt]
Agent $\to$ \texttt{go south} $\to$ WM: ``$-$= Living Room $=$$-$'' \quad {\color{gray}(navigates away from stove)}\\
Agent $\to$ \texttt{go north} $\to$ WM: ``You can't go that way.'' \quad {\color{red!70!black}(broken room connectivity)}\\
Steps 10--50: Kitchen $\leftrightarrow$ Living Room navigation loop; ``\texttt{look}'' in Living Room returns Bedroom description.
}}
}}
\caption{\textbf{Complete trajectory comparison on TextWorld} (\texttt{textworld\_169}, Qwen3-8B agent, LLaMA-8B WM). Steps 1--7 are functionally identical across conditions. At the critical decision point (step 8), the BehR-WM preserves correct game-state semantics, including error handling for invalid verbs, enabling task completion. The W2W-WM exhibits broken room connectivity that traps the agent in a navigation loop.}
\label{fig:full_trajectory}
\end{figure*}

\section{Controlled Perturbation Experiments}
\label{sec:appendix_perturbation}

This section provides full seven-perturbation results for the non-judge automated metrics summarized in \S\ref{sec:reward_validation}, together with definitions for all metrics used in the diagnostic comparison.

\subsection{Metric Definitions}
\label{sec:appendix_metric_definitions}

All metrics compare a perturbed candidate observation \(\tilde{s}_{t}\) with the original environment observation \(s_{t}\) from the same transition. Scores are averaged over the same perturbation set, and higher values indicate closer agreement with the original transition.

\begin{description}
\item[BehR.] BehR scores whether the perturbed observation preserves the Reference Agent's support for the logged next action \(a^*_{t+1}\). We compute the Reference-Agent log-likelihood of \(a^*_{t+1}\) under both \(s_t\) and \(\tilde{s}_t\), then use the likelihood-gap reward in Eq.~\ref{eq:behr}. For the perturbation tables only, we report a monotone normalized score in \([0,1]\) so that all metrics share the same higher-is-better presentation scale. Because the underlying likelihood gap is real-valued for every candidate observation, it can serve as a dense reward for GRPO training.
\item[ACS.] ACS~\citep{guan2026computer} is an action-consistency score: a frozen agent chooses actions under the original and perturbed observations, and the score is one when the decoded actions match and zero otherwise. We average this binary score over samples. ACS is action-sensitive at evaluation time, but the discrete match signal is not a dense reward in our setting.
\item[GPT-4o Judge.] The content-only GPT-4o judge follows the LLM-as-a-judge paradigm~\citep{zheng2023judging}: GPT-4o compares the original and perturbed observations without seeing the logged next action, and returns a normalized semantic/content-preservation score. It can therefore reward textual plausibility even when the perturbation removes decision-critical information.
\item[F1 Reward.] F1 Reward follows the surface-overlap reward family used by RLVR-World~\citep{wu2026rlvr}, which defines an F1 reward over predicted and ground-truth changed items. In our perturbation study, we instantiate the same F1 principle over tokens in the original and perturbed observations. This baseline measures lexical overlap and ignores whether the preserved words support the same next action.
\item[BERTScore and ROUGE-L.] BERTScore~\citep{zhang2019bertscore} measures contextual semantic similarity, while ROUGE-L~\citep{lin-2004-rouge} measures longest-common-subsequence overlap. Both are standard text-similarity metrics; neither conditions on the agent action.
\item[Exact Match.] Exact match is one only when the perturbed observation exactly equals the original observation, and zero otherwise. It is a surface-form metric and gives no graded signal for non-identical perturbations.
\end{description}

\subsection{8B Reference-Agent Results}

\begin{table*}[tp]
\centering
\small
\setlength{\tabcolsep}{4pt}
\resizebox{\textwidth}{!}{%
\begin{tabular}{llccccccc}
\csname toprule\endcsname
{\bfseries Perturbation} & {\bfseries Severity} & {\bfseries N} & {\bfseries BehR} & {\bfseries ACS} & {\bfseries F1} & {\bfseries BERT} & {\bfseries ROUGE-L} & {\bfseries EM} \\
\midrule
Oracle & None & 100 & \textbf{0.996} & 0.740 & 1.000 & 1.000 & 1.000 & 100\% \\
Shuffle & Mild & 100 & 0.671 & 0.660 & 1.000 & 0.953 & 0.726 & 0\% \\
Drop Irrelevant & Mild & 100 & 0.763 & 0.600 & 0.803 & 0.859 & 0.803 & 0\% \\
Add Irrelevant & Moderate & 100 & 0.845 & 0.740 & 0.891 & 0.892 & 0.891 & 0\% \\
Random Noise & Moderate & 100 & 0.174 & 0.090 & 0.576 & 0.666 & 0.297 & 0\% \\
\rowcolor{red!8} \textbf{Drop Target} & \textbf{Severe} & 100 & \textbf{0.100} & 0.370 & \textbf{0.961} & \textbf{0.974} & \textbf{0.961} & 0\% \\
Random Cross & Severe & 100 & 0.091 & 0.160 & 0.598 & 0.763 & 0.511 & 0\% \\
\bottomrule
\end{tabular}
}
\caption{\textbf{Controlled perturbations with an 8B Reference Agent.} Scores over $N=100$ shared samples; behavior-sensitive metrics should rank Drop~Target below Drop~Irrelevant. Surface metrics invert this ordering despite the decision-critical target removal. ACS follows \citet{guan2026computer}; F1, BERT, ROUGE-L, and EM denote F1 Reward~\citep{wu2026rlvr}, BERTScore~\citep{zhang2019bertscore}, ROUGE-L~\citep{lin-2004-rouge}, and Exact Match.}
\label{tab:perturbation}
\end{table*}

Content-only LLM judges fail for the same reason as BERTScore: they compare semantic resemblance rather than action consequences.
Action-conditioned judging can partially recover the right ordering, but it either depends on unavailable oracle information or yields subjective discrete ratings.
The main text therefore focuses on a compact DI-vs.-DT comparison, while the full seven-perturbation table above provides the controlled validation for the non-judge automated metrics.

\subsection{Reward Design Alternatives}
\label{sec:appendix_reward_design_alternatives}

During reward design, we also compared logged-action BehR with two richer action-conditioned alternatives. \textbf{Multi-action BehR} aggregates likelihood gaps over the logged action and additional admissible candidate actions, while \textbf{ACS} measures decoded action agreement between original and perturbed states. Table~\ref{tab:reward_design_alternatives} reports the raw diagnostic scores on controlled perturbations.

\begin{table}[ht]
\centering
\small
\setlength{\tabcolsep}{4pt}
\begin{tabular*}{\columnwidth}{l@{\extracolsep{\fill}}ccc}
\csname toprule\endcsname
{\bfseries Condition} & {\bfseries Logged BehR} & {\bfseries Multi BehR} & {\bfseries ACS} \\
\midrule
Oracle          & 0.895 & 0.892 & 0.606 \\
Drop Irrelevant & 0.763 & 0.210 & 0.600 \\
Drop Target     & 0.100 & 0.291 & 0.370 \\
Random Cross    & 0.299 & 0.255 & 0.277 \\
Random Noise    & 0.299 & 0.252 & 0.332 \\
\bottomrule
\end{tabular*}
\caption{\textbf{Controlled comparison of reward-design alternatives.} A useful reward should preserve a high score on Oracle states and assign lower scores to behaviorally degraded states. Logged-action BehR best separates Drop~Target from Drop~Irrelevant.}
\label{tab:reward_design_alternatives}
\end{table}

Multi-action aggregation under-penalizes Drop~Target because averaging over multiple admissible actions dilutes the loss of support for the logged decision. It also over-penalizes Drop~Irrelevant, scoring it below Drop~Target under its own scale. Logged-action BehR avoids this inversion because it probes the action whose validity is established by the trajectory. It is also computationally lighter: scoring additional candidate actions would require additional Reference-Agent continuations and would make the reward more sensitive to the reference policy's preferences among plausible actions.

\section{Trajectory-Level Analysis}
\label{sec:appendix_trajectory}
\label{sec:appendix_webshop_analysis}

We present detailed trajectory-level analyses that support the calibrated-surrogate findings in \S\ref{sec:surrogate_eval}.

\begin{table}[ht]
\centering
\small
\setlength{\tabcolsep}{6pt}
\begin{tabular*}{\columnwidth}{l@{\extracolsep{\fill}}rrr}
\toprule
\textbf{Agent} & \textbf{Real} & \textbf{W2W} & \textbf{BehR} \\
\midrule
\multicolumn{4}{l}{\textit{WebShop}} \\
\midrule
Qwen3-0.6B  & 6.5  & 4.5  & 4.0  \\
Qwen3-8B    & 14.5 & 16.5 & 17.0 \\
Qwen3-32B   & 16.5 & 19.0 & 15.5 \\
GPT-4o      & 19.0 & 19.0 & 21.0 \\
GPT-5       & 39.0 & 39.0 & 43.0 \\
\midrule
\multicolumn{4}{l}{\textit{TextWorld}} \\
\midrule
Qwen3-0.6B  & 0.0   & 42.5  & 9.5   \\
Qwen3-1.7B  & 3.5   & 37.5  & 19.5  \\
Qwen3-8B    & 87.0  & 100.0 & 97.0  \\
Qwen3-32B   & 97.0  & 100.0 & 99.5 \\
GPT-5       & 100.0 & 100.0 & 99.5 \\
\bottomrule
\end{tabular*}
\caption{\textbf{Surrogate success rates across agents.} Success rates in the real environment and in W2W/BehR WMs; BehR reduces weak-agent overestimation on TextWorld while preserving overall ranking.}
\label{tab:surrogate_eval}
\end{table}

\paragraph{Episode Length and Timeout.}
Table~\ref{tab:episode_length} reports mean episode length and max-step timeout rates across conditions.
In real WebShop, episodes average 8--17 steps, with \emph{zero} timeouts.
In WM-based evaluation, 82--97\% of episodes reach the 50-step limit: agents enter exploratory loops because the WM cannot faithfully reproduce WebShop's purchase-completion signal.
TextWorld shows no such effect: strong agents (14B) maintain nearly identical episode lengths across all conditions ($\sim$12 steps), confirming that the TextWorld WM preserves the task's temporal structure.

\begin{table}[ht]
\centering
\small
\setlength{\tabcolsep}{3pt}
\begin{tabular*}{\columnwidth}{l@{\extracolsep{\fill}}lrrrr}
\toprule
 & & \multicolumn{2}{c}{\textbf{WebShop}} & \multicolumn{2}{c}{\textbf{TextWorld}} \\
\cmidrule(lr){3-4} \cmidrule(lr){5-6}
\textbf{Agent} & \textbf{Cond.} & \textbf{Mean} & \textbf{T/O\%} & \textbf{Mean} & \textbf{T/O\%} \\
\midrule
\multirow{3}{*}{0.6B} & Real & 8.5 & 0\% & 49.2 & 100\% \\
 & W2W-WM & 47.8 & 94.5\% & 36.5 & 57.5\% \\
 & BehR-WM & 48.0 & 94.0\% & 46.2 & 90.5\% \\
\midrule
\multirow{3}{*}{1.7B} & Real & 12.6 & 0\% & 48.6 & 96.5\% \\
 & W2W-WM & 43.8 & 85.6\% & 37.1 & 62.5\% \\
 & BehR-WM & 42.4 & 82.2\% & 42.9 & 80.5\% \\
\midrule
\multirow{3}{*}{4B} & Real & 16.6 & 0\% & 21.5 & 25.5\% \\
 & W2W-WM & 48.2 & 96.0\% & 14.5 & 8.5\% \\
 & BehR-WM & 48.9 & 97.0\% & 17.0 & 15.5\% \\
\midrule
\multirow{3}{*}{14B} & Real & 17.3 & 0\% & 11.9 & 1.5\% \\
 & W2W-WM & 46.7 & 91.5\% & 11.0 & 0\% \\
 & BehR-WM & 46.5 & 91.3\% & 11.8 & 2.5\% \\
\bottomrule
\end{tabular*}
\caption{\textbf{Episode length and timeout rates.} Mean steps and percentage of episodes reaching the 50-step limit in real and WM environments.}
\label{tab:episode_length}
\end{table}

\paragraph{Failure modes.}
WebShop failures look qualitatively different in Real and in WM-based evaluation.
In the real environment, weaker agents most often buy the wrong item (72.2\% for 0.6B; 38.9\% for 1.7B), while stronger agents more often stop before purchasing (74.3\% for 4B and 74.6\% for 14B).
In WM-based evaluation, failures are overwhelmingly timeouts: 98.9--100\% for W2W-WM and 99.0--100\% for BehR-WM across the same agents.
This structural difference explains why WM-internal SR is not directly comparable to real SR and motivates W2R replay as the definitive evaluation protocol.

\paragraph{Task-level agreement.}
BehR-WM improves WebShop agreement for every Qwen evaluation agent we tested, from 91.0\% to 92.5\% (0.6B), 81.2\% to 84.4\% (1.7B), 92.4\% to 92.9\% (4B), 77.8\% to 81.0\% (8B), 90.3\% to 91.3\% (14B), and 81.4\% to 83.9\% (32B).
Unlike TextWorld, where the dominant gap between W2W-WM and BehR-WM is concentrated in false positives for weak agents, WebShop shows a more mixed FP/FN pattern; still, false positives fall or stay flat in every configuration, including 7 to 5 (0.6B), 17 to 12 (8B), and 21 to 15 (32B).
The consistency across scales complements the TextWorld result in Table~\ref{tab:task_agreement} and supports the claim that BehR improves calibration rather than merely shifting one operating point.

\paragraph{Episode Length Calibration on TextWorld.}
On TextWorld, where the timeout pathology does not confound episode structure, BehR-WM produces mean episode lengths consistently closer to the real environment than W2W-WM.
Across all four agent--backbone configurations, the mean step gap $|\bar{T}_{\text{WM}} - \bar{T}_{\text{Real}}|$ is smaller for BehR-WM:
Qwen3-8B with Qwen-WM: $1.6 \to 0.8$ steps;
Qwen3-8B with LLaMA-WM: $5.1 \to 2.3$ steps;
Qwen3-32B with Qwen-WM: $0.6 \to 0.4$ steps;
Qwen3-32B with LLaMA-WM: $4.7 \to 1.5$ steps.
The effect is largest on the LLaMA-base WM, where W2W-WM inflates mean episode length by 37--42\% relative to Real; BehR-WM reduces this to 14--16\%.
We do not report per-task step analysis on WebShop, where 82--97\% of WM episodes reach the 50-step limit regardless of WM variant (Table~\ref{tab:episode_length}), rendering step-level comparisons uninformative.

\FloatBarrier

\section{Lookahead Planning Results}
\label{sec:appendix_lookahead_main}

Table~\ref{tab:lookahead} in the main text reports the completed lookahead planning results on WebShop ($K{=}5$).
This appendix section clarifies how to interpret those results alongside the paired tests in Appendix~\ref{sec:appendix_ci}.

Both W2W-WM and BehR-WM improve over ReAct, showing that world-model rollouts can provide useful short-horizon decision support. Under the same $K{=}5$ planning protocol, BehR-WM further improves over W2W-WM for both Qwen3-8B and GPT-4o, suggesting that more behaviorally aligned predicted futures translate into better planning decisions. The paired McNemar tests below evaluate the end-to-end gain of BehR-WM lookahead over no-planning ReAct on the same WebShop tasks.

\section{From Functional Consistency to Practical Proxy}
\label{sec:appendix_proxy}

Functional consistency (Eq.~\ref{eq:fe}) requires preserving the agent's full action distribution $\pi(\cdot \mid \hat{s}_t) \approx \pi(\cdot \mid s_t)$, whereas BehR (Eq.~\ref{eq:behr}) monitors only the likelihood assigned by a frozen Reference Agent $\pi_{\text{ref}}$ to the next action $a^*_{t+1}$ recorded in the offline trajectory, rather than the downstream agent $\pi$ itself.
The gap is twofold: \emph{what} is preserved (one action vs.\ the full distribution) and \emph{who} measures it (Reference Agent vs.\ downstream agent).

\paragraph{Why this relaxation works.}
In our environments, the critical behavioral question at each step is often close to binary: does the agent still select the correct action?
Preserving the logged expert action's likelihood can therefore approximate the decision boundary between the correct action and the most relevant alternatives.
The controlled perturbation study (Table~\ref{tab:perturbation}) supports this interpretation: BehR assigns 0.100 to Drop~Target (optimal action destroyed) vs.\ 0.763 to Drop~Irrelevant (optimal action preserved), suggesting that the single-action proxy captures a task-relevant part of the decision boundary.

\paragraph{Limitations.}
The proxy is weakest when multiple plausible actions exist and behavior depends on their relative ranking rather than just the top choice.
It is also mediated by a specific frozen Reference Agent (Qwen3-8B); the cross-agent results in Table~\ref{tab:main_cr} provide partial reassurance, but transfer is heterogeneous.

\section{Reference-Agent Family Overlap}
\label{sec:appendix_judge_confound}
\label{sec:appendix_cross_ref}

Table~\ref{tab:cross_ref_main} reports a direct Reference-Agent ablation on WebShop with Ministral-3-8B and LLaMA3.1-8B Reference Agents.

A natural concern is that BehR gains are inflated when the evaluated agent belongs to the same family as the Qwen3-8B Reference Agent.
For the Qwen-base WM, same-family agents gain $+0.064$ on average versus $+0.023$ for cross-family, but the gap shrinks to $+0.064$ vs.\ $+0.045$ after excluding TextWorld rows where GPT agents already hit the CR$_{\text{pw}} \geq 0.990$ ceiling.
The LLaMA-base WM provides the cleanest test: Reference Agent (Qwen3-8B), WM backbone (LLaMA3.1-8B), and evaluated agents (GPT-4o/5) are three distinct families.
Cross-family agents average $\Delta\text{CR}_{\text{pw}} = +0.078$, \emph{exceeding} same-family $+0.040$, directly contradicting the confounding hypothesis.

\section{Statistical Analysis}
\label{sec:appendix_ci}

We assess the aggregate direction of the task-level results across the 16 domain--backbone--agent cells in Table~\ref{tab:main_cr}. Our main claim is not that every row-level improvement is independently significant, but that the direction of the effect is consistent across the full benchmark suite.

\paragraph{Sign tests across benchmark cells.}
Table~\ref{tab:sign_tests} reports pairwise sign tests over the 16 domain--backbone--agent cells. Ties are excluded from the binomial test. BehR-WM improves over W2W-WM, F1-WM, and SurR-WM in 13 cells with no losses.

\begin{table}[ht]
\centering
\small
\setlength{\tabcolsep}{7pt}
\begin{tabular*}{\columnwidth}{l@{\extracolsep{\fill}}cccc}
\csname toprule\endcsname
{\bfseries Comparison} & {\bfseries Wins} & {\bfseries Ties} & {\bfseries Losses} & {\bfseries $p$} \\
\midrule
BehR vs. W2W  & 13 & 3 & 0 & $2.44\times10^{-4}$ \\
BehR vs. F1   & 13 & 3 & 0 & $2.44\times10^{-4}$ \\
BehR vs. SurR & 13 & 3 & 0 & $2.44\times10^{-4}$ \\
\bottomrule
\end{tabular*}
\caption{\textbf{Suite-level sign tests.} Pairwise comparisons over the 16 CR$_{\text{pw}}$ cells; ties are excluded from the two-sided binomial test.}
\label{tab:sign_tests}
\end{table}

\paragraph{Rank analysis.}
As a complementary omnibus test, we rank W2W-WM, F1-WM, SurR-WM, and BehR-WM within each of the 16 cells, using mid-ranks for ties. A Friedman rank test rejects equal method ranks ($\chi^2=23.32$, $p=3.5\times10^{-5}$). Mean ranks are BehR-WM $1.28$, SurR-WM $2.69$, F1-WM $2.94$, and W2W-WM $3.09$, with lower being better. The Nemenyi critical difference is approximately $1.17$ at $\alpha=0.05$, so BehR-WM is separated from all three baselines under this suite-level rank comparison.

\paragraph{Paired bootstrap over cells.}
We also resample the 16 cells with replacement $B=10{,}000$ times and compute the mean CR$_{\text{pw}}$ difference between BehR-WM and each baseline. Table~\ref{tab:bootstrap_ci} reports the resulting mean differences, percentile confidence intervals, and one-sided bootstrap $p$-values.

\begin{table}[ht]
\centering
\small
\setlength{\tabcolsep}{4pt}
\begin{tabular*}{\columnwidth}{l@{\extracolsep{\fill}}ccc}
\csname toprule\endcsname
{\bfseries Comparison} & {\bfseries Mean $\Delta$} & {\bfseries 95\% CI} & {\bfseries $p$} \\
\midrule
BehR $-$ W2W  & $+0.051$ & $[+0.030, +0.077]$ & $<10^{-4}$ \\
BehR $-$ F1   & $+0.055$ & $[+0.033, +0.080]$ & $<10^{-4}$ \\
BehR $-$ SurR & $+0.044$ & $[+0.020, +0.074]$ & $<10^{-4}$ \\
\bottomrule
\end{tabular*}
\caption{\textbf{Paired bootstrap over benchmark cells.} Mean CR$_{\text{pw}}$ differences between BehR-WM and each baseline over $B=10{,}000$ bootstrap resamples of the 16 cells.}
\label{tab:bootstrap_ci}
\end{table}

\paragraph{Lookahead McNemar tests.}
For downstream planning, we compare BehR-WM lookahead against no-planning ReAct using paired McNemar tests on the same 200 WebShop tasks. Table~\ref{tab:mcnemar_lookahead} reports the discordant-pair counts. The gains over ReAct are statistically detectable for the two agents shown, complementing the main-text comparison where BehR-WM also improves over W2W-WM under the same lookahead protocol.

\begin{table}[ht]
\centering
\small
\setlength{\tabcolsep}{4pt}
\begin{tabular*}{\columnwidth}{l@{\extracolsep{\fill}}cccc}
\csname toprule\endcsname
{\bfseries Agent} & {\bfseries LA only} & {\bfseries ReAct only} & {\bfseries $\chi^2$} & {\bfseries $p$} \\
\midrule
Qwen3-8B & 34 & 14 & 7.52 & 0.006 \\
GPT-4o   & 31 & 16 & 4.17 & 0.041 \\
\bottomrule
\end{tabular*}
\caption{\textbf{Paired McNemar tests for lookahead.} LA-only wins are tasks solved by BehR-WM lookahead but not ReAct; ReAct-only wins are tasks solved by ReAct but not BehR-WM lookahead.}
\label{tab:mcnemar_lookahead}
\end{table}

These analyses provide suite-level evidence for the aggregate direction of the results, while larger task sets and broader planning evaluations remain important future work.